\documentclass[lettersize,journal]{IEEEtran}
\usepackage{amsmath,amsfonts,amssymb}
\usepackage{algorithmic}
\usepackage{algorithm}
\usepackage{array}
\usepackage{subcaption}
\usepackage{textcomp}
\usepackage{stfloats}
\usepackage{url}
\usepackage{verbatim}
\usepackage{graphicx}
\usepackage{booktabs}
\usepackage{multirow}
\usepackage{cite}
\newcommand{\methodname}{IGDPR}
\newcommand{\tivae}{TI-VAE}

\providecommand{\Description}[1]{}

\newcommand{\xspace}{\mathcal{X}}
\newcommand{\yspace}{\mathcal{Y}}

\newcommand{\xdim}{d}

\newcommand{\dtrain}{\mathcal{D}_{\text{train}}}
\newcommand{\dval}{\mathcal{D}_{\text{val}}}
\newcommand{\dtest}{\mathcal{D}_{\text{test}}}
\newcommand{\dsyn}{\mathcal{D}_{\text{syn}}}
\newcommand{\daug}{\mathcal{D}_{\text{aug}}}

\newcommand{\xreal}{x}
\newcommand{\yreal}{y}
\newcommand{\xtilde}{\tilde{x}}
\newcommand{\ytilde}{\tilde{y}}
\newcommand{\zlat}{z}

\newcommand{\nreal}{N}
\newcommand{\nsyn}{M}
\newcommand{\vaeenc}{e}
\newcommand{\vaedec}{d}
\newcommand{\gen}{\mathcal{G}}

\newcommand{\weightfunc}{w}
\newcommand{\loss}{\ell}
\newcommand{\obj}{\mathcal{L}}
\newcommand{\pred}{f}
\newcommand{\predparams}{{\psi^{(p)}}}

\newcommand{\predmodel}{\pred_{\predparams}}
\newcommand{\E}{\mathbb{E}}

\begin{document}

\title{Rethinking Data Augmentation under Covariate Shift:\\ Invariant-Guided Diffusion and Prototype Reweighting}

\author{
Hongyu Cao, Xinyuan Wang, Arun Vignesh Malarkkan, Kunpeng Liu, Haifeng Chen and Yanjie Fu%
\thanks{Hongyu Cao, Xinyuan Wang, Arun Vignesh Malarkkan and Yanjie Fu are with the
School of Computing and Augmented Intelligence, Arizona State University,
Tempe, AZ 85281 USA (e-mail: hongyu.jack.cao@gmail.com, xwang735@asu.edu, arun.malarkkan@asu.edu, yanjie.fu@asu.edu).}%
\thanks{Kunpeng Liu is with the School of Computing, Clemson University,
Clemson, SC 29634 USA (e-mail: kunpenl@clemson.edu).}%
\thanks{Haifeng Chen is with the Data Science and System Security Department,
NEC Laboratories America, Princeton, NJ 08540 USA (e-mail: haifeng@nec-labs.com).}%
}

\markboth{IEEE Transactions on Knowledge and Data Engineering}%
{Anonymous \MakeLowercase{\textit{et al.}}: Invariant-Guided Diffusion and Prototype Reweighting}

\maketitle

\begin{abstract}
In many industrial applications,
1) tabular data is scarce and imbalanced and thus requires synthetic expansion;
2) input distributions drift between training and deployment (covariate shift);
3) validation sets often diverge from unseen test environments; or
4) standard generative models simply mimic outdated source distributions.
This learning setting limits the stability of standard augmentation and adaptation pipelines. We generalize the task under such setting as the Augmented and Weighted Learning under Covariate Shift problem (AWL-CS).
AWL-CS imposes two critical challenges on existing methods:
1) misleading generative guidance where models optimize for source similarity rather than downstream task relevance, and
2) structural instability of distributional density where reweighting mechanisms overfit to noisy validation signals.
To tackle these challenges, we propose \methodname{} (Invariant-Guided Diffusion with Prototype Reweighting), a unified framework that synergizes stable synthesis and structural adaptation:
i) To achieve task-relevant generation, we steer the diffusion sampling process using invariant potentials to ensure synthetic samples align with stable decision boundaries rather than outdated correlations.
ii) To ensure stable adaptation, we develop a prototype-based reweighting strategy that assesses sample reliability through structural clusters instead of isolated points, effectively filtering validation noise. Extensive experiments on real data demonstrate our method improves data quality by augmenting the most beneficial data for robust learning.
\end{abstract}


\section{Introduction}
\label{sec:intro}
\begin{figure}[!t]
  \centering
  \includegraphics[width=\linewidth]{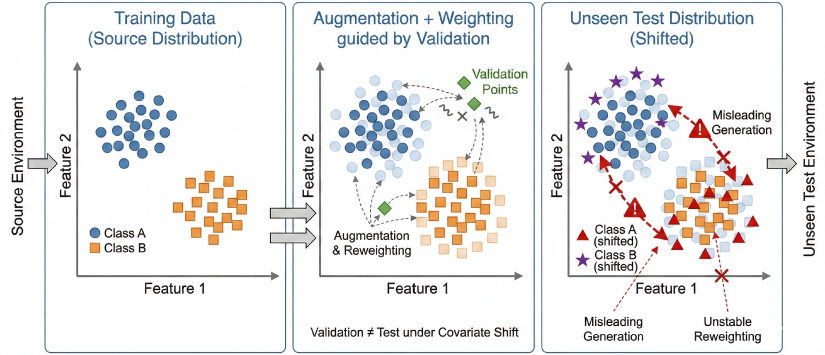}
\caption{Motivation of Augmented and Weighted Learning under Covariate Shift (AWL-CS).
Under distribution shift, validation-guided augmentation and reweighting may amplify spurious patterns instead of improving generalization.}
\label{fig:motivation}
\end{figure}

\IEEEPARstart{M}{any} industrial applications, such as healthcare analytics and fraud detection, rely heavily on tabular data for high-stakes decision-making~\cite{stoian2026surveydeeplearningapproaches, VALLEVIK2024105413, b2026diffusiondrivensynthetictabulardata, golec2025interpretablellmscreditrisk, choi2017generating, liu2026continuousoptimizationfeatureselection, liu2026permutationinvariantrepresentationlearningrobust}.
Generative augmentation is deployed to address data scarcity and class imbalance~\cite{xu2019modeling, liu2025climbclassimbalancedlearningbenchmark, herurkar2025tabulardataadaptersimproving, Zhang2023Sensitive}. However, real-world environments are dynamic; input distributions often drift between training and deployment (e.g., evolving fraud tactics or shifting patient demographics), thus, result into covariate shift~\cite{shimodaira2000improving, quinonero2009dataset}.
Under this context, the validation set doesn't represent the unseen test environment, making standard augmentation and reweighting ineffective.
This scenario can be generalized as a new learning problem: \textit{\textbf{A}ugmented and \textbf{W}eighted \textbf{L}earning under \textbf{C}ovariate \textbf{S}hift (AWL-CS)} (\textbf{Figure \ref{fig:motivation}}).
AWL-CS enables a model to robustly generalize by leveraging invariant-guided synthesis and structure-aware reweighting. Solving AWL-CS can address multiple critical issues to increase the reliability of AI systems. For example, in many scenarios, 1) data is scarce and imbalanced, thus synthetic expansion is needed; 2) distribution shifts make  gradients of source or training data misleading for future tasks; 3) validation sets contain noises that destabilizes traditional reweighting; or 4) models must prioritize stable, invariant patterns to ensure safe generalization in unseen environments.

There are two major challenges in solving AWL-CS:
1) misleading generative guidance under shift, and
2) structural instability in reweighting.
First, traditional generative augmentation focuses on maximizing statistical similarity to the source training data.
However, under covariate shift, the source distribution diverges from the unseen target environment.
Therefore, the generator relies on outdated source gradients, thus, produce samples that fit the past, instead of optimizing for the downstream task. Misleading generative guidance seeks to answer: how can we steer the data generation process to synthesize task-relevant samples when reliable target gradients are unavailable?
Second, reweighting mechanisms suffer from structural instability of distributional density and coverage.
This is because:
1) these methods align the data distribution to a validation set, thus, fail to represent the shifting test environment; and
2) they estimate weights for isolated points, are easily influenced by validation noises.
A weighting mechanism that aligns with noisy validation signals will destabilize the predictor.
Therefore, structural instability in reweighting is intended to answer: how can we assign reliable importance scores that prioritize stable structural clusters over spurious validation artifacts?

In short, existing studies show the inability of any single line of work to jointly address misleading generative guidance and structural instability under shift; a more detailed comparison is deferred to Section~\ref{sec:related}.

\textbf{Our insights: teaming invariant generation and reweighing.} We formulate a generic problem of augmented and weighted learning for covariate-shifted, label-scarce, and structurally unstable tabular environments.
We show that AWL can be solved by synergizing invariant-guided synthesis and prototype-based reweighting, where the synthesis creates task-aligned samples to fill data gaps, and the reweighting assigns reliability scores to filter noise. In particular:
i) Stable guidance is critical for effective augmentation. We find that leveraging invariant potentials to steer the diffusion process can fight against the bias introduced by outdated source gradients for robust data expansion.
ii) We demonstrate that to correct distribution shifts without absorbing validation artifacts, the weighting mechanism needs to be structure-aware rather than point-wise. This requirement can be reformulated into a task of prototype-based reliability assessment within an invariant latent space. We show that the reliability assessment is indeed a function of trajectory stability and local density relative to cluster centroids.
Solving this structural alignment effectively helps the model to achieve both high-fidelity data augmentation and stable risk minimization in unseen test environments.

\textbf{Summary of Proposed Approach}. Inspired by these findings, we present \methodname{} (Invariant-Guided Diffusion with Prototype Reweighting), a principled and generic augmentation-adaptation framework for the AWL-CS problem by synergizing invariant-guided synthesis and prototype-based reweighting.
The framework has two goals: 1) generating task-aligned samples that resist evolving distribution shifts; 2) ensuring stable adaptation by mitigating validation noise.
To achieve Goal \#1, we develop an Invariant-Guided Diffusion model. In particular, we find that generative utility can be improved by constraining the diffusion process with invariant structure knowledge. We design an invariant potential function that modifies the reverse diffusion sampling, effectively steering the generated data away from outdated source correlations and towards the stable decision boundary of the target environment. To achieve Goal \#2, we develop a Prototype-Based Reweighting strategy. In particular, we reformulate the instability problem into a structural matching task. We propose to assess sample reliability using prototype clusters rather than isolated points. By assigning weights based on the density of reliable prototypes in the invariant space, we effectively minimize the variance caused by noisy validation signals.
Finally, we present extensive empirical results to demonstrate the effectiveness of our method in boosting performance across various real-world tabular datasets under covariate shifts.

\section{Related Work}
\label{sec:related}

AWL-CS sits at the intersection of three established lines of research, none of which jointly resolves the misleading-guidance and structural-instability challenges raised in Section~\ref{sec:intro}. We review each line and highlight the specific gap our method targets.

\subsection{Generative Data Augmentation}
Generative models such as GANs and diffusion models~\cite{xu2019modeling, xu2018tgan, kotelnikov2023tabddpm, song2021scorebased, ho2020ddpm, kingma2013auto} are widely used to enlarge tabular training sets. Conditional GANs (CTGAN, TVAE) use mode-specific normalization to address multimodality~\cite{xu2019modeling}, while diffusion variants like TabDDPM~\cite{kotelnikov2023tabddpm} achieve broader mode coverage through iterative refinement. However, the dominant objective in this line---maximizing fidelity to the static source distribution~\cite{cheng2024downstreamtaskorientedgenerativemodel, MUMUNI2022100258}---is precisely what fails under shift: source gradients are outdated proxies for an unseen target environment~\cite{borisov2021deep}, so high-fidelity samples can amplify spurious correlations rather than fill genuine gaps~\cite{11122661}. Recent work on guided diffusion explores classifier or energy guidance for image domains~\cite{shiftresilient}, but tabular adaptations rarely incorporate invariance signals, leaving an open question on how to steer the denoising trajectory when target supervision is unavailable.

\subsection{Distribution Reweighting under Covariate Shift}
Importance weighting and density-ratio estimation methods~\cite{shimodaira2000improving, sugiyama2007covariate, huang2007correcting, sugiyama2007kliep} correct covariate shift by re-scaling training samples toward a validation reference~\cite{nguyen2023generalregularizationcovariateshift, zhou2023modelagnosticsamplereweighting}. The fundamental assumption is that the validation set is a faithful proxy for the test distribution---a premise that breaks under genuine covariate shift, where validation itself diverges from $\dtest$. As a result, point-wise weight estimates inherit validation noise and destabilize training~\cite{cao2026flatconsensusdiffusionrobustdata}. Related lines on noisy-label learning~\cite{jiang2018mentornet, han2018coteaching, cao2026sim2actrobustsimulationtodecisionlearning} and group-robust optimization~\cite{sagawa2020groupdro} provide partial remedies but operate at the instance or pre-defined group level, missing the structural cluster signal that our prototype-based reweighting exploits.

\subsection{Invariant and Prototype Representation Learning}
Invariant risk minimization~\cite{arjovsky2019invariant, cao2026mitigatingshortcutreasoninglanguage}, domain adversarial training~\cite{ganin2016dann}, and gradient-alignment methods~\cite{shi2021gradient} learn representations whose risk is stable across environments. Prototype-based approaches such as Prototypical Networks~\cite{snell2017prototypical}, PTaRL~\cite{ye2024ptarl}, and prototype matching~\cite{lin2021prototype} provide complementary metric-space structure~\cite{liu2026hierarchicalpermutationinvariantfeaturetransformation}. These methods are effective at aligning representations but remain decoupled from generative mechanisms: they do not synthesize new samples to cover sparse regions, so the underlying scarcity that motivates augmentation in the first place is left unaddressed.

\subsection{Positioning of \methodname}
Our framework unifies the three perspectives above. From generative augmentation, we inherit a latent diffusion backbone for broad mode coverage; from invariant learning, we inherit an environment-invariance constraint that yields stable supervision when validation is unreliable; and from prototype-based methods, we inherit a structural-cluster view that aggregates noisy point-wise signals into reliable reweighting. Crucially, these components are coupled rather than stacked: invariant potentials directly steer the diffusion sampling trajectory (rather than being applied as a post-hoc filter), and prototype reweighting operates inside the same invariant latent space rather than in raw feature space. This coupling is what enables \methodname{} to simultaneously fill data gaps with task-aligned samples and assign reliability scores that resist validation noise.

\section{Problem Statement}
\label{sec:problem}

\textbf{The AWL-CS Problem.}
We consider a supervised learning setup with input $\xspace \in \mathbb{R}^{\xdim}$, label $\yspace$, and a latent environment variable $E \in \mathcal{E}$ indexing operating conditions, time periods, or population sub-strata. Training data is a mixture of source environments $\dtrain = \bigcup_{e \in \mathcal{E}_{\text{train}}} \mathcal{D}_e$, while test samples may originate from environments $E' \in \mathcal{E}_{\text{test}}$ unobserved at training time. We assume \emph{covariate shift with invariant conditional}, $\mathcal{P}_{\text{train}}(Y\!\mid\!X) = \mathcal{P}_{\text{test}}(Y\!\mid\!X)$, while the marginal $\mathcal{P}(X)$ drifts across environments; since $E$ is unobserved at test time, predictions must be stable across $\mathcal{E}$.
We address this problem via a four-stage pipeline: partition, augmentation, reweighing, and optimization.

\noindent\underline{\emph{Partition:}}
The dataset is partitioned into three disjoint subsets: a training set $\dtrain$ for parameter updates, a validation set $\dval$ for guiding hyper-parameters, and an unseen testing set $\dtest$ for final evaluation.
In operational environments, such as high-stakes industrial analytics, data is often scarce and the distribution shifts between the collection phase ($\dtrain \cup \dval$) and the deployment phase ($\dtest$).

\noindent\underline{\emph{Augmentation:}}
A generator $\gen$ uses observed data to create auxiliary synthetic samples $\dsyn$. These are combined with real data to form the augmented dataset:
\begin{equation}
    \daug = \dtrain \cup \dsyn = \{(\xreal_i, \yreal_i)\}_{i=1}^{\nreal} \cup \{(\xtilde_j, \ytilde_j)\}_{j=1}^{\nsyn},
    \label{eq:augmented_set}
\end{equation}
where $\nreal$ and $\nsyn$ denote the numbers of real and synthetic samples, respectively.

\noindent\underline{\emph{Reweighing:}} Subsequently, a weighting function $\weightfunc(\xreal, \yreal)$ assigns a scalar importance score to each pair in $\daug$ to shape the distribution.
However, a core assumption mismatch exists in this standard pipeline: it typically assumes that $\dval$ is a sufficient proxy for $\dtest$. In realistic covariate shift settings, $\dval$ often diverges from the test environment.
The consequence is that the generator mimics outdated patterns and the weighting function overfits to validation noise, leading to suboptimal generalization.

\noindent\underline{\emph{Optimization:}}
The optimization objective is to construct a framework to i) synthesize task-relevant samples that bridge the distributional gap, and ii) assign stable importance weights to filter noise and ensure generalization.
Formally, our goal is to learn a predictor $\predmodel: \xspace \to \yspace$ that minimizes the expected risk on the unseen target distribution:
\begin{equation}
    \predparams^* = \mathop{\arg\min}_{\predparams} \E_{(\xreal, \yreal) \sim \dtest} [\loss(\predmodel(\xreal), \yreal)].
    \label{eq:generalization_goal}
\end{equation}
Existing methods cannot be directly applied because $\dtest$ is unavailable during training to guide the optimization.
Therefore, the practical learning objective is to approximate the target risk by minimizing the weighted empirical risk on the augmented dataset:
\begin{equation}
    \mathcal{L}_{\text{final}}(\predparams) = \sum_{(\xreal, \yreal) \in \daug} \weightfunc(\xreal, \yreal) \cdot \loss(\predmodel(\xreal), \yreal),
    \label{eq:weighted_objective}
\end{equation}
A successful solution must derive weights $\weightfunc(\xreal,\yreal)$ and synthetic samples $\dsyn$ such that maximizing performance on Eq.~\eqref{eq:weighted_objective} translates to maximizing performance on Eq.~\eqref{eq:generalization_goal}, effectively bridging the distribution shift.

\section{The \methodname{} Framework}

\begin{figure*}[!t]
    \centering
    \includegraphics[width=\linewidth]{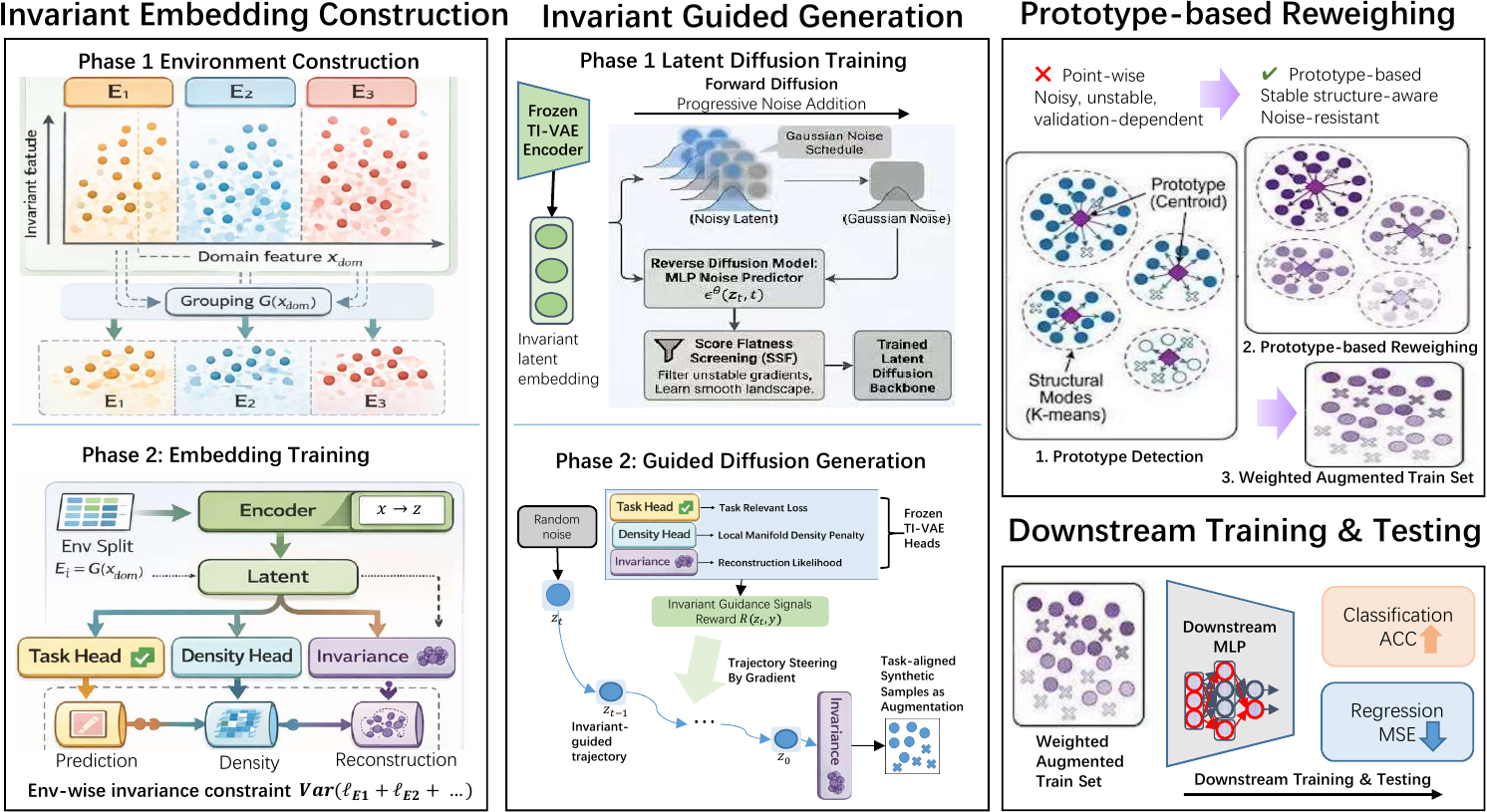}
    \caption{Overview of the proposed \methodname{} (Invariant-Guided Diffusion with Prototype Reweighting) framework.
    The pipeline consists of four tightly coupled stages.
    (1) Invariant embedding construction: training data are partitioned into multiple pseudo-environments based on domain features, and a task-invariant latent space is learned via a multi-head \tivae{} with task, density, and invariance supervision.
    (2) Invariant-guided latent diffusion generation: a latent diffusion model is trained in the invariant space and, during sampling, its denoising trajectories are actively steered by frozen invariant guidance signals toward task-aligned and structurally stable regions.
    (3) Prototype-based reweighting: real and synthetic samples are clustered into structural prototypes, where reliability scores are estimated at the prototype level to avoid noisy and unstable point-wise reweighting.
    (4) Weighted downstream training: the downstream predictor is trained on the weighted augmented dataset to improve generalization under covariate shift.}
    \label{fig:pipeline}
\end{figure*}

\subsection{Overview}
Figure~\ref{fig:pipeline} shows the framework includes three integrated components:
1) \textbf{Invariant-Guided Generation.} To synthesize task-relevant data, we avoid generating directly in the noisy raw space. Instead, we employ a Task-Invariant VAE (\tivae) to project inputs into a disentangled latent space. Within this space, a diffusion backbone synthesizes new representations, strictly guided by invariant potentials to preserve semantic consistency.
2) \textbf{Prototype-Based Reweighting.} Recognizing that generative quality varies, we introduce a reweighting mechanism to quantify sample reliability. After decoding latents back to the data space, we assess their structural alignment with learned prototypes, assigning a scalar weight to every synthetic sample to downweight outliers.
3) \textbf{Augmented Boosting.} Finally, we construct a robust downstream predictor. By uniting original real data with these weighted synthetic samples, we optimize the predictor to leverage expanded diversity while suppressing unreliable augmentations.

\subsection{Task-Invariant Training Data Embedding}
\label{sec:ti_vae}

\noindent\textbf{Why Task-Invariant VAE Matters?}
Real world shifts and noises can lead to nonalignment between explicit training, validation, and testing data.
Representation learning provides an opportunity to construct a latent space that disentangles stable task semantics from shifting domain noise, in order to enable effective generation and reweighting under shift.
However, standard autoencoders typically focus solely on reconstruction minimization, thus, the learned embedding space often fails to eliminate domain-specific patterns and erroneous correlations found in the source distribution.
In addition to reconstruction signal, which ensures the semantic validity of the decoded synthetic data, embedding learning must be supervised by two additional signals:
1) \textit{invariant task signal} that provides stable supervision to steer data generation away from outdated source correlations;
2) \textit{density signal} that reveals the underlying distribution structure to support prototype-based reliability assessment.

\noindent\textbf{Modeling Intuition: Disentanglement via Invariance.}
Our intuition relies on the principle of invariant risk minimization~\cite{arjovsky2019invariant}. If a latent representation captures the true structure of the task, it should be able to predict the label consistently across distinct environments, regardless of the distributional shift.
By partitioning the training data into pseudo-environments (i.e., groups of data points) and enforcing the variance of the prediction risk to be zero across them, we compel the encoder to strip away environment-sensitive features (noise) and converge on an invariant subspace. This ensures that the subsequent diffusion model learns to generate samples of core data distribution instead of superficial variations.

\noindent\textbf{\tivae{} Embedding Model.}
We adopt a VAE backbone to encode training samples into latent vectors, comprising: i) an encoder $\vaeenc$ that maps each training point $\mathbf{\xreal} \in \mathbb{R}^d$ to variational parameters (mean $\vec{\mu}$ and variance $\vec{\sigma}$), from which the embedding $\mathbf{\zlat} \in \mathbb{R}^k$ is drawn as a Gaussian sample;
ii) a decoder $\vaedec$ that reconstructs $\mathbf{\hat{\xreal}}$ from $\mathbf{\zlat}$, given by:
\begin{equation}
    (\vec{\mu}, \vec{\sigma}) = \vaeenc(\mathbf{\xreal}), \quad \mathbf{\zlat} \sim \mathcal{N}(\vec{\mu}, \text{diag}(\vec{\sigma}^2)), \quad \hat{\mathbf{\xreal}} = \vaedec(\zlat).
    \label{eq:enc_dec}
\end{equation}
We attach two auxiliary projection heads to the embedding:
i) a task head $\hat{y} = f(\mathbf{\zlat})$ to estimate class labels ($y$);
ii) a density head $\hat{d} = h(\mathbf{\zlat})$ to estimate local manifold density (i.e., how crowded is the neighborhood around this point), measured by the inverse average distance to its $K$-nearest neighbors in the latent space.

\noindent\textbf{Multiple Pseudo-environment Construction to Enforce Task Invariance.}
To enforce invariance constraint, we first define  multiple distinct environments (groups of data points) within training data.
In particular, we divide training data features $\mathbf{x}$ into
i) invariant features  ($\mathbf{x}_{\mathrm{inv}}$);
ii) domain-specific features ($\mathbf{x}_{\mathrm{dom}}$). We partition the training set into $K$ pseudo-environments using a grouping function $\Pi: \mathbb{R}^{\xdim_{\mathrm{dom}}} \to \{1,\dots,K\}$ that maps each sample's domain features to an environment index:
\begin{equation}
\begin{aligned}
    \mathbf{x} &= (\mathbf{x}_{\mathrm{inv}}, \mathbf{x}_{\mathrm{dom}}), \\
    E_i &= \{(\mathbf{x}, y) \in \dtrain :
    \Pi(\mathbf{x}_{\mathrm{dom}}) = i\}, \\
    &\qquad i=1,\dots,K.
\end{aligned}
\label{eq:env_const}
\end{equation}
Whether $\mathbf{x}_{\mathrm{dom}}$ is provided directly (e.g., explicit operating conditions on C-MAPSS) or must be recovered automatically (e.g., on Scania APS, where no explicit domain attributes exist) determines whether $\Pi$ is a trivial lookup of the labeled attribute or a learned clustering on a surrogate feature; per-dataset choices are reported in Section~\ref{sec:setup}.
In this way, we can factorize the joint distribution to ensure that the label dependence relies solely on the invariant features (i.e., $p(y\mid \mathbf{x}) \approx p(y\mid \mathbf{x}_{\mathrm{inv}})$). Such partitioning can penalize representations that rely on the domain-specific features.
\noindent\textit{Construction in two phases.}
The grouping function $\Pi$ is realized in two phases. \emph{Phase 1 (Initial Partitioning):} we run $K$-Means on the domain-specific subspace, assigning each sample to its nearest centroid $E^{(0)}(i) = \arg\min_{k} \|\mathbf{x}_{\mathrm{dom}}^{(i)} - c_k\|_2^2$. \emph{Phase 2 (Adaptive Merging):} to ensure each environment carries enough samples for a stable invariance penalty, we set a minimum-size threshold $\tau$ and iteratively merge the smallest environment $E_{\text{small}}$ into its centroid-nearest neighbor $j^* = \arg\min_{j \neq i} \|\mu(E_i) - \mu(E_j)\|_2$, where $\mu(E)$ is the mean of $\mathbf{x}_{\mathrm{dom}}$ within $E$. The procedure is summarized in Algorithm~\ref{alg:env_gen}.

\begin{algorithm}[h]
    \caption{Pseudo-environment Construction with Adaptive Merging}
    \label{alg:env_gen}
    \begin{algorithmic}[1]
        \REQUIRE Training set $\mathcal{D} = \{(\mathbf{x}_i, y_i)\}$, domain features $\{\mathbf{x}_{\mathrm{dom}}^{(i)}\}$, initial clusters $K$, minimum size threshold $\tau$
        \STATE \textbf{Initialize:} Apply $K$-Means on $\{\mathbf{x}_{\mathrm{dom}}^{(i)}\}$ to obtain partitions $\mathcal{E} = \{E_1, \dots, E_K\}$
        \STATE Compute centroids $\mu_k$ for each $E_k \in \mathcal{E}$
        \WHILE{$\min_{E \in \mathcal{E}} |E| < \tau$}
            \STATE $E_{\text{small}} \leftarrow \arg\min_{E \in \mathcal{E}} |E|$
            \STATE $E_{\text{target}} \leftarrow \arg\min_{E \in \mathcal{E}, E \neq E_{\text{small}}} \|\mu(E_{\text{small}}) - \mu(E)\|_2$
            \STATE \textbf{Merge:} $E_{\text{target}} \leftarrow E_{\text{target}} \cup E_{\text{small}}$; remove $E_{\text{small}}$ from $\mathcal{E}$
            \STATE Recompute $\mu(E_{\text{target}})$
        \ENDWHILE
        \RETURN Final environments $\mathcal{E}$
    \end{algorithmic}
\end{algorithm}

\noindent\textbf{Task-Invariant Objective.}
The optimization objective integrates the reconstruction ($\mathcal{L}_{\mathrm{VAE}}$), task ($\mathcal{L}_{\text{task}}$), density ($\mathcal{L}_{\text{den}}$), and invariance losses:
\begin{equation}
    \mathcal{L} = \mathcal{L}_{\mathrm{VAE}} + \lambda_{\text{task}} \cdot \mathcal{L}_{\text{task}} + \lambda_{\text{den}} \cdot \mathcal{L}_{\text{den}} + \lambda_{\text{inv}} \cdot \mathcal{L}_{\text{inv}},
    \label{eq:total_loss}
\end{equation}
where $\lambda_{\text{task}}, \lambda_{\text{den}}, \lambda_{\text{inv}}$ are the loss weights reported in Table~\ref{tab:hyperparams}, and the invariance loss is the cross-environment risk variance
$
\mathcal{L}_{\text{inv}} = \mathrm{Var}_{E_i^K} ( \mathbb{E}_{x \in \mathcal{D}_e}\,\ell_{\text{task}}(f(z), y) )
$.

The reconstruction term is the (negative) evidence lower bound, kept separate from $\mathcal{L}_{\text{task}}$ so that the prior regularization is not double-counted with task supervision:
$
    \mathcal{L}_{\mathrm{VAE}} = \mathbb{E}_{x} \Big[ \mathbb{E}_{q_\phi} [\log p_\theta(x\mid z)] - \beta_{\mathrm{KL}}\, D_{\mathrm{KL}}(q_\phi(z\mid x)\,\|\,p(z)) \Big]
$.
The task and density losses are to minimize the classification or regression error and the mean squared error against log-probability density of density estimation respectively:
$
    \mathcal{L}_{\mathrm{task}} = \mathbb{E}_{(x,y)} [\ell_{\text{task}}(f(z),y)], \quad \mathcal{L}_{\mathrm{den}} = \mathbb{E}_{x} [\| h(z) - \log p_{\text{train}}(x) \|_2^2]
$.
Lastly, we introduce the term $\mathrm{Var}_{E_i^K} ( \mathbb{E}_{x \in \mathcal{D}_e}\,\ell_{\text{task}}(f(z), y) )$ to represent the variance of the risk across the constructed environments.
By minimizing this variance, we penalize embedding that fluctuate due to domain shifts, effectively forcing the encoder to extract stable, invariant patterns.

\subsection{Invariant-Guided Diffusion Generation}

\noindent\textbf{Why Latent Invariant Guided Diffusion Matters?}
To synthesize data, we can use standard generative models (GANs, vanilla diffusion) that maximizes the likelihood of training data.
However, under the contexts of shifts, strictly mimicking training distribution is harmful; it learns unaligned correlations that do not generalize to the unseen test environment.
To move beyond simple imitation of training data, we need a generation mechanism that synthesizes samples adhering to stable, task-invariant patterns.

\noindent\textbf{Modeling Intuition: Navigation in Invariant Space.}
We treat data generation as a directed navigation problem within a latent manifold:
\tivae{} learns the distribution landscape of the invariant training data embeddings;
the diffusion generator should learn how to navigate the landscape to sample more data points to represent the underlying generalizable distribution.
Without a guided direction, the diffusion generator would simply wander back to resample the regions of seen training data.
To prevent this, we introduce a direction guidance signal: an invariant reward.
During the generation of new samples, this reward steers the denoising trajectory away from regions of drifted patterns and toward regions of high task relevance and structural stability under the density lens.

\noindent\textbf{Step 1: Learning Data Distribution Landscape via Diffusion.}
The first step is to learn data distribution.
We employ a Multi-Layer Perceptron (MLP) as the backbone for a latent noise prediction network.
The training objective is to learn the reverse transition probability, allowing the model to reconstruct coherent latent structures from pure noise.
The process consists of a forward diffusion stage and a backward denoising stage.
In the forward process, we progressively corrupt a clean latent code $\zlat_0$ into a noisy state $\zlat_t$ by adding Gaussian noise $\epsilon$.
This follows the Markovian transition $q(\zlat_t \mid \zlat_0) = \mathcal{N}(\zlat_t; \sqrt{\bar{\alpha}_t}\zlat_0, (1-\bar{\alpha}_t)I)$, where $\alpha_t$ is the ratio of original signal to preserve and $I$ is the identity matrix.
In the backward process, the MLP network learns to predict the added noise to approximate the reverse transition $p(\zlat_{t-1} \mid \zlat_t)$, optimized by the standard DDPM noise-prediction loss $\mathbb{E}_{t,\zlat_0,\epsilon}\bigl[\|\epsilon - \hat{\epsilon}_\omega(\zlat_t,t)\|_2^2\bigr]$.

\noindent\textbf{Step 2: Guided Denoising as Data Generation.}
After learning the generic data distribution, we aim to generate task-aligned samples using guided denoising.
We propose to guide the denoising (sampling) process using signals from the frozen \tivae{} model (Section~\ref{sec:ti_vae}).

\noindent\textit{How invariance enters the guidance.} Note that the reward $\mathcal{R}$ defined below does \emph{not} contain a separate invariance term. Instead, invariance enters \emph{implicitly} through the frozen \tivae{}: (i) the latent state $\zlat_t$ itself lives in the invariant subspace learned by the encoder under the cross-environment variance penalty in Eq.~\eqref{eq:total_loss}; (ii) the task head $f_\psi$ in the reward is the invariant task head trained jointly with that penalty, so the task term scores how well an invariant predictor agrees with the target label; (iii) the density head $h_\zeta$ and decoder $d$ are likewise trained on the invariant representation, so their gradients only respond to variation along invariant axes. As a result, every gradient term in Eq.~\eqref{eq:grad_update} is computed inside the invariant geometry, and guidance pushes trajectories along invariant directions even though $\mathcal{R}$ itself only aggregates task, density, and reconstruction.

We define a guidance reward function $\mathcal{R}$ that aggregates these three signals as:
\begin{equation}
    \mathcal{R}(\zlat_t, \tilde{y}) = \underbrace{-\ell_{\text{task}}(f_\psi(\hat{\zlat}_0), \tilde{y})}_{\text{Task}} - \underbrace{\mu_{\text{den}} \cdot h_\zeta(\hat{\zlat}_0)}_{\text{Density}} + \underbrace{\log p_{\text{data}}(d(\hat{\zlat}_0))}_{\text{Reconstruction}},
    \label{eq:noise_reward}
\end{equation}
where $\hat{\zlat}_0$ is the estimated clean latent embedding derived from $\zlat_t$ via Tweedie's formula~\cite{Efron2011TweediesFA}, $\mu_{\text{den}}>0$ is a reward-side density weight (distinct from the training-time $\lambda_{\text{den}}$ in Eq.~\eqref{eq:total_loss}), and $p_{\text{data}}$ denotes a fixed marginal data-density estimator fit on $\dtrain$ (e.g., a kernel density or VAE-decoder likelihood) so that the reconstruction term scores the realism of the decoded sample. The task term ensures the sample belongs to the target class $\tilde{y}$ and the density term avoids over-dense regions.

\textit{2.1) Steering the Denoising (Sampling) Trajectory.}
During the reverse denoising steps, we modify the standard DDPM update rule by injecting the gradient of the reward function to steer the generation towards the invariant decision boundary:
\begin{equation}
    \zlat_{t-1} = \frac{1}{\sqrt{\alpha_t}}\!\left(\zlat_t - \frac{1-\alpha_t}{\sqrt{1-\bar{\alpha}_t}}\,\hat{\epsilon}_\omega(\zlat_t,t)\right) + \gamma_t \cdot \nabla_{\zlat_t}\log \mathcal{R}(\zlat_t, \tilde{y}),
    \label{eq:grad_update}
\end{equation}
where the bracketed term is the standard DDPM denoising mean that moves $\zlat_t$ toward its predicted clean state, and $\gamma_t$ is the guidance strength that controls how strongly we enforce invariance.
Each denoising iteration/trajectory can generate an embedding of a high-quality data point to synthesize.

\textit{2.2) Constructing Synthesized Data from Decoding.}
To construct synthetic dataset, we perform $M$ denoising iterations and obtain a set of optimized latent embeddings, denoted by $\mathcal{Z}_{syn}$.
We decode these data embeddings into a set of explicit data points using the pretrained decoder $d(\cdot)$ and develop a synthetic dataset:
$\mathcal{D}_{syn} = \left\{ (d(\tilde{z}_i), \tilde{y}_i) \mid (\tilde{z}_i, \tilde{y}_i) \in \mathcal{Z}_{syn} \right\}, i\in [1, M]$.
The synthetic dataset is not merely a variation of the source data, but a re-imagined distribution optimized for stability and task performance under shift.

\noindent\textbf{Practical Notes.}
We adopt four implementation choices to keep guided sampling stable and efficient:
\begin{itemize}
  \item \textit{Log-gradient scaling.} We apply a temperature calibration $R\leftarrow \exp(R/\tau)$ together with norm clipping on $\|\nabla\log R\|$ to prevent excessively large updates that would push trajectories off the latent manifold.
  \item \textit{Computationally efficient gradients.} When $R_{\text{inv}}$ and $R_{\text{task}}$ are functions of a low-dimensional latent $\phi(\zlat_t)$, we compute $\nabla_{\zlat_t}\log R = J_{\phi}(\zlat_t)^\top \nabla_{\phi}\log R$ via the chain rule, where $J_{\phi}$ is the Jacobian of $\phi$. This keeps guidance cheap whenever $\phi$ has small dimension.
  \item \textit{Avoiding joint training leakage.} Guidance signals are applied \emph{post-training} on a frozen \tivae, preventing the diffusion backbone from collapsing onto trivial solutions that artificially maximize $R$ while ignoring fidelity.
  \item \textit{Score estimation in guidance.} When the score $\nabla_{\zlat_t}\log p_\theta$ used inside $\nabla_{\zlat_t}\log\mathcal{R}$ is costly to evaluate directly, we substitute a proxy---either the norm of the denoiser output $\hat{\epsilon}_\omega(\zlat_t,t)$ or a finite-difference estimate---which preserves the gradient direction at a fraction of the cost.
\end{itemize}

\subsection{Efficient Prototype-based Reweighing of Augmented Data}
\label{sec:proto_reweight}

\noindent\textbf{Why Prototype-Based Reweighting Matters?}
Because invariant-guided diffusion improves coverage, but reweighting is still needed to correct residual distribution mismatch and prevent synthetic samples from distorting the effective training distribution.
Standard reweighting methods optimize weights for individual samples, which makes them highly susceptible to overfitting the specific noise patterns within the small validation set.
Our intuition is to enforce a structural constraint: a sample should be trusted only if its local neighborhood is collectively stable and task-aligned. By aggregating reliability signals such as task loss and invariance consistency at the prototype level, we can effectively filter out high-frequency validation noise.

\noindent\textbf{Modeling Intuition: Assigning Weights to Mixture of Structural Modes.}
We treat the augmented data as a collection of structural modes.
Instead of treating each sample as an independent entity, we group samples that share similar characteristics in the raw data feature space.
We identify cluster centers as prototypes that represent the dominant patterns of the underlying distribution.
We calculate weights at the prototype level to create a robust basis for data quality assessment and ignore individual sample-level noise.

\noindent\textbf{Step 1: Prototype Identification in the Invariant Latent Space.}
The first step is to partition the augmented dataset $\mathcal{D}_{aug} = \mathcal{D}_{train} \cup \mathcal{D}_{syn}$ into clusters.
We cluster on the invariant latent representations $Z = \Phi(X)$ produced by the frozen \tivae{} encoder, so that prototype structure aligns with task-relevant geometry rather than nuisance variation in raw features.
We apply K-Means to divide the augmented latents into $K$ distinct groups.
The centroids of these groups are defined as prototypes, representing the structural modes of the augmented distribution under the invariant geometry.

\noindent\textbf{Step 2: Prototype-level Reweighting.}
We weigh each cluster by considering: i) the task and invariance losses and ii) its neighborhood density.
The intuition is to assign higher weights to regions that are structurally stable (low prediction and invariance errors) while down-weighting over-represented regions.
Specifically, we aggregate the sample-level task, invariance, and density losses within each cluster $k$ to form the cluster-level weights.
The prototype weight $\alpha_k$ is formulated as:
\begin{equation}
    \alpha_k = \frac{1}{\left( {\ell}_{\text{task}}^k + {\ell}_{\text{inv}}^k \right)^{\beta} \cdot \left( {\ell}_{\text{den}}^k \right)^{\gamma}}
    \label{eq:proto_importance}
\end{equation}
where ${\ell}_{\text{task}}^k$ and ${\ell}_{\text{inv}}^k$ are the average task cross-entropy loss and invariance loss for samples in the cluster $k$, representing the cluster's instability.
${\ell}_{\text{den}}^k$ represents the local density magnitude (or latent density estimation where higher values indicate higher density).
The hyperparameters $\beta > 0$ and $\gamma > 0$ control the sensitivity of the weighting mechanism: a larger $\beta$ penalizes unstable clusters more heavily, while a larger $\gamma$ imposes stronger penalties on redundant, high-density regions.

\noindent\textbf{Step 3: Individual Sample Weight Assignment.}
Step 3 is to assign weights to individual samples based on their parent prototypes.
We assign the prototype weight equally to each data point, so the weight of the $i$-th individual sample is given by $w_i = \frac{\alpha_{k}}{n_{k}}$,
where $k$ denotes the nearest cluster index of the $i$-th sample, $\alpha_{k}$ is the importance of the $k$-th prototype, and $n_{k}$ is the total number of samples in that cluster.
The same weight for cluster is for computational efficiency since calculating loss for every data point to assign weight will be time-consuming.

\begin{table}[!t]
    \centering
    \caption{Summary of Datasets and Industrial Tasks.}
    \label{tab:datasets}
    \begin{tabular}{lcccc}
        \toprule
        \textbf{Dataset} & \textbf{Source} & \textbf{Task Type} & \textbf{Samples} & \textbf{Feat Dim.} \\
        \midrule
        \textbf{C-MAPSS} & Simulated & Regression & $\sim$53k & 14 \\
        \textbf{Scania APS} & Real & Classification & 76k & 170 \\
        \textbf{Gas Sensor} & Real & Classification & 13k & 128 \\
        \textbf{MetroPT-3} & Real & Regression & 1.5M & $\sim$5 \\
        \bottomrule
    \end{tabular}
\end{table}

\begin{table*}[!t]
\centering
\caption{\textbf{Main Results on Industrial Domain Shift Benchmarks.} We compare different generative augmentation methods combined with various reweighting strategies. We report \textbf{Accuracy}, \textbf{F1-Macro} (F1-M), and \textbf{F1-Weighted} (F1-W) for classification tasks, and \textbf{MSE}, \textbf{RMSE}, and \textbf{$R^2$} for regression tasks. The \textbf{Guided Latent Diffusion} generator denotes our invariant-guided latent diffusion module; paired with the \textbf{Prototype} reweighter (bold row) it constitutes our full method \methodname{}. \textbf{Bold} indicates best performance.}
\label{tab:main_results}
\resizebox{\textwidth}{!}{%
\begin{tabular}{ll ccc ccc ccc ccc}
\toprule
\multirow{2}{*}{\textbf{Generator}} & \multirow{2}{*}{\textbf{Reweighter}} &
\multicolumn{3}{c}{\textbf{Gas Sensor} (Classif.)} &
\multicolumn{3}{c}{\textbf{NASA C-MAPSS} (Reg.)} &
\multicolumn{3}{c}{\textbf{Scania ComponentX} (Classif.)} &
\multicolumn{3}{c}{\textbf{MetroPT-3} (Reg.)} \\
\cmidrule(lr){3-5} \cmidrule(lr){6-8} \cmidrule(lr){9-11} \cmidrule(lr){12-14}
 & &
Acc $\uparrow$ & F1-M $\uparrow$ & F1-W $\uparrow$ &
MSE $\downarrow$ & RMSE $\downarrow$ & $R^2$ $\uparrow$ &
Acc $\uparrow$ & F1-M $\uparrow$ & F1-W $\uparrow$ &
MSE $\downarrow$ & RMSE $\downarrow$ & $R^2$ $\uparrow$ \\
\midrule
\multirow{4}{*}{\textbf{No Aug}}
 & Uniform    & 0.6479 & 0.6066 & 0.6069 & 2069.21 & 45.49 & 0.82 & 0.9083 & 0.8710 & 0.9021 & 0.2605 & 0.5104 & 0.71 \\
 & Importance & 0.6600 & 0.6225 & 0.6542 & 2067.63 & 45.47 & 0.83 & 0.9104 & 0.8748 & 0.9046 & 0.2589 & 0.5088 & 0.72 \\
 & Approx.    & 0.4414 & 0.3921 & 0.4356 & 2093.73 & 45.76 & 0.78 & 0.8897 & 0.8415 & 0.8823 & 0.2516 & 0.5016 & 0.73 \\
 & Prototype  & 0.6615 & 0.6284 & 0.6571 & 2064.41 & 45.44 & 0.83 & 0.9146 & 0.8793 & 0.9085 & 0.2598 & 0.5097 & 0.72 \\
\midrule
\multirow{4}{*}{\textbf{MLP GAN}}
 & Uniform    & 0.6410 & 0.6032 & 0.6354 & 2117.05 & 46.01 & 0.75 & 0.9121 & 0.8769 & 0.9068 & 0.2521 & 0.5021 & 0.73 \\
 & Importance & 0.6511 & 0.6117 & 0.6453 & 2120.64 & 46.05 & 0.74 & 0.9132 & 0.8785 & 0.9079 & 0.2514 & 0.5014 & 0.73 \\
 & Approx.    & 0.4202 & 0.3814 & 0.4136 & 2113.66 & 45.97 & 0.75 & 0.8925 & 0.8453 & 0.8854 & 0.2568 & 0.5067 & 0.72 \\
 & Prototype  & 0.6450 & 0.6095 & 0.6408 & 2059.92 & 45.38 & 0.84 & 0.9188 & 0.8849 & 0.9132 & 0.2536 & 0.5036 & 0.73 \\
\midrule
\multirow{4}{*}{\textbf{CT GAN}}
 & Uniform    & 0.6526 & 0.6178 & 0.6472 & 2054.08 & 45.33 & 0.85 & 0.9145 & 0.8802 & 0.9089 & 0.2558 & 0.5058 & 0.72 \\
 & Importance & 0.6375 & 0.6019 & 0.6324 & 2085.61 & 45.67 & 0.79 & 0.9128 & 0.8776 & 0.9071 & 0.2540 & 0.5040 & 0.72 \\
 & Approx.    & 0.4174 & 0.3789 & 0.4102 & 2088.98 & 45.71 & 0.78 & 0.8951 & 0.8487 & 0.8879 & 0.2595 & 0.5094 & 0.71 \\
 & Prototype  & 0.6509 & 0.6162 & 0.6458 & 2088.85 & 45.71 & 0.78 & 0.9201 & 0.8873 & 0.9152 & 0.2569 & 0.5068 & 0.72 \\
\midrule
\multirow{4}{*}{\textbf{TabDDPM}}
 & Uniform    & 0.6652 & 0.6318 & 0.6594 & 2059.16 & 45.38 & 0.84 & 0.9162 & 0.8821 & 0.9105 & 0.2419 & 0.4918 & 0.75 \\
 & Importance & 0.6541 & 0.6203 & 0.6487 & 2075.56 & 45.57 & 0.81 & 0.9151 & 0.8804 & 0.9092 & 0.2398 & 0.4897 & 0.76 \\
 & Approx.    & 0.4287 & 0.3920 & 0.4211 & 2077.77 & 45.60 & 0.80 & 0.8986 & 0.8539 & 0.8921 & 0.2425 & 0.4924 & 0.74 \\
 & Prototype  & 0.6724 & 0.6396 & 0.6668 & 2051.91 & 45.32 & 0.85 & 0.9220 & 0.8891 & 0.9178 & 0.2367 & 0.4866 & 0.77 \\
\midrule
\multirow{4}{*}{\shortstack{\textbf{Guided}\\\textbf{Latent}\\\textbf{Diffusion}}}
 & Uniform    & 0.6796 & 0.6465 & 0.6751 & 2051.88 & 45.32 & 0.86 & 0.9185 & 0.8846 & 0.9134 & 0.2462 & 0.4962 & 0.74 \\
 & Importance & 0.6236 & 0.5892 & 0.6181 & 2087.02 & 45.69 & 0.79 & 0.9169 & 0.8827 & 0.9118 & 0.2437 & 0.4937 & 0.75 \\
 & Approx.    & 0.4336 & 0.3958 & 0.4269 & 2081.64 & 45.62 & 0.80 & 0.9012 & 0.8563 & 0.8945 & 0.2454 & 0.4954 & 0.74 \\
 & Prototype  & \textbf{0.6909} & \textbf{0.6587} & \textbf{0.6864} & \textbf{2044.63} & \textbf{45.22} & \textbf{0.87} & \textbf{0.9262} & \textbf{0.8948} & \textbf{0.9217} & \textbf{0.2305} & \textbf{0.4801} & \textbf{0.78} \\
\bottomrule
\end{tabular}%
}
\end{table*}

\subsection{Boosting Prediction via Integrated New Weights and Augmented Data}
With the augmented data and corresponding new weights, we can retrain a boosted predictive model.
Without loss of generality, let us use a MLP as the down stream predictive model $g$ trained on the augmented dataset $\daug$, where the $i$-th data point is associated with a weight $\weightfunc_i$.
The downstream MLP training objective is given by:
\begin{equation}
    \obj(g) = \sum_{(\mathbf{x}, y, \weightfunc) \in \daug} \weightfunc \cdot \loss\big(g(\mathbf{x}), y\big)
    \label{eq:down_loss}
\end{equation}
This weighted and augmented learning strategy allows the decision boundary to be shaped by trustworthy patterns from both the real and synthetic domains.

\subsection{Theoretical Justification}
\label{sec:theory}

We now connect the prototype reweighter, the density penalty, and invariance-guided synthesis to established results in importance-weighted risk minimization and information-theoretic domain generalization.

\noindent\textbf{Setup.}
Adopting the notation of Section~\ref{sec:problem}, let $\predmodel:\xspace\to\yspace$ be a predictor with bounded loss $\loss\in[0,1]$. The ideal target risk is
\begin{equation}
\mathcal{R}_{\text{target}}(\theta) = \E_{(X,Y)\sim \mathcal{P}_{\text{target}}}\big[\loss(\predmodel(X),Y)\big],
\end{equation}
but only $\mathcal{P}_{\text{train}}$ is observed. Under covariate shift with invariant label conditional ($\mathcal{P}_{\text{target}}(Y\!\mid\!X)\!=\!\mathcal{P}_{\text{train}}(Y\!\mid\!X)$), the optimal importance weight is $w^*(X)=\mathcal{P}_{\text{target}}(X)/\mathcal{P}_{\text{train}}(X)$. Our prototype reweighter approximates this ratio inside an invariant latent space $Z=\Phi(X)$, where $\Phi$ denotes the frozen \tivae{} encoder map (distinct from the downstream predictor $g$ in Eq.~\eqref{eq:down_loss}), upweighting underrepresented prototypes rather than estimating point-wise ratios on noisy validation data.

\noindent\textbf{Density penalty and effective sample size.}
We analyze a continuous relaxation of the prototype weights in Eq.~\eqref{eq:proto_importance}. Writing $d_{\text{proto}}(z)$ for the soft distance to the nearest prototype and noting that $-\log\ell_{\text{task}}^k$ and $-\log\ell_{\text{den}}^k$ are the log-domain counterparts of the factors in $\alpha_k$, we work with
\begin{equation}
w(z) \;\propto\; \exp\!\bigl(-\beta\,d_{\text{proto}}(z) - \gamma \log p_{\text{train}}(z)\bigr),
\label{eq:w_relax}
\end{equation}
which preserves the qualitative behaviour of $\alpha_k$ (upweighting stable, underrepresented regions) while being amenable to standard importance-weighting analysis. The corresponding effective sample size,
\begin{equation}
n_{\text{eff}} = \frac{(\sum_i w_i)^2}{\sum_i w_i^2},
\label{eq:n_eff}
\end{equation}
governs the generalization gap of weighted ERM. Substituting the sample-to-cluster mapping $w_i = \alpha_k/n_k$ from Section~\ref{sec:proto_reweight} (with $n_k$ the cluster size) yields the closed cluster-form expression
\begin{equation}
n_{\text{eff}} = \frac{(\sum_k \alpha_k)^2}{\sum_k \alpha_k^2 / n_k}.
\label{eq:n_eff_cluster}
\end{equation}
Increasing $\gamma$ in Eq.~\eqref{eq:proto_importance} shrinks $\alpha_k$ for high-density clusters, flattening the denominator $\sum_k \alpha_k^2/n_k$ while leaving $(\sum_k \alpha_k)^2$ approximately constant---directly inflating $n_{\text{eff}}$ and tightening the bound in Prop.~1.

\noindent\textbf{Proposition 1 (Weighted ERM bound).}
\textit{Assume (A1) the loss is bounded, $\loss\in[0,1]$; (A2) the weights $\{w_i\}$ in Eq.~\eqref{eq:w_relax} are bounded above by a constant $W<\infty$ and approximate the density ratio $w^*(X)=\mathcal{P}_{\text{target}}(X)/\mathcal{P}_{\text{train}}(X)$ in the sense that $\E_{\mathcal{P}_{\text{train}}}[(w-w^*)^2]$ is small; (A3) samples $\{(X_i,Y_i)\}_{i=1}^n$ are drawn i.i.d.\ from $\mathcal{P}_{\text{train}}$; and (A4) the hypothesis class has bounded complexity (e.g., finite Rademacher complexity).} Then, for any $\theta$ in the hypothesis class and any $\delta\in(0,1)$, with probability at least $1-\delta$,
\begin{equation}
\mathcal{R}_{\text{target}}(\theta) \;\le\; \hat{\mathcal{R}}_w(\theta) + C \cdot \sqrt{\frac{\log(1/\delta)}{n_{\text{eff}}}} + \varepsilon_w,
\label{eq:prop1_bound}
\end{equation}
where the constant $C$ depends on $W$ and the hypothesis-class complexity, and $\varepsilon_w$ collects the approximation error of $w$ relative to $w^*$. The bound follows from a Bernstein-type concentration inequality on the importance-weighted empirical process (cf.~\cite{shimodaira2000improving, sugiyama2007covariate}); a detailed argument is standard and omitted.

Thus, the density factor $(\ell_{\text{den}}^k)^{\gamma}$ in Eq.~\eqref{eq:proto_importance} inflates $n_{\text{eff}}$ by penalizing redundant high-density modes, which tightens the second term of Eq.~\eqref{eq:prop1_bound}.

\noindent\textit{Remark (relating $w(z)$ to $\alpha_k$).} The form in Eq.~\eqref{eq:w_relax} is a continuous, log-domain relaxation of $\alpha_k$: cluster-level reliability $({\ell}_{\text{task}}^k + {\ell}_{\text{inv}}^k)^{-\beta}$ corresponds to the exponential of $-\beta\,d_{\text{proto}}$ when the task/invariance loss is interpreted as a soft cluster-distance, and the density factor $({\ell}_{\text{den}}^k)^{-\gamma}$ corresponds to $\exp(-\gamma\log p_{\text{train}})$. The \emph{cluster-to-sample bridge} is exact and given by Eq.~\eqref{eq:n_eff_cluster}; the remaining \emph{continuous relaxation} from $\alpha_k$ to $w(z)$ (e.g., via local Lipschitz arguments) is left to future work.

\noindent\textbf{Proposition 2 (Approximate Information Bottleneck view).}
\textit{Assume (B1) the encoder $g$ has sufficient capacity to realize the conditional risk $\E_x[\loss(f(z),y)]$ as a function of $z$; (B2) the per-environment empirical risks $\{\E_{x\in\mathcal{D}_e}\loss_{\text{task}}(f(z),y)\}_{e=1}^K$ converge to their population counterparts; and (B3) the cross-environment variance term in Eq.~\eqref{eq:total_loss} is driven to zero.} Then the \tivae{} objective acts as a tractable surrogate for the constrained problem
\begin{equation}
\min_{g} I(Z;E) \quad \text{s.t.}\quad I(Z;Y) \ge c,
\label{eq:approx_ib}
\end{equation}
in the sense that vanishing risk variance implies the conditional risk $\E[\loss\mid Z, E]$ no longer depends on $E$, which under (B1)--(B2) approximates the conditional independence $Y\perp E\mid Z$ and therefore
\begin{equation}
\mathcal{P}(Y\mid Z, E') \approx \mathcal{P}(Y\mid Z),
\end{equation}
for any unseen environment $E'$. We emphasize that this is a relaxation: the variance penalty equates the \emph{expected loss} across environments, not the full conditional distribution, so the gap between Eq.~\eqref{eq:approx_ib} and the implemented objective is non-zero in general. A formal connection between risk-variance penalties and $I(Z;E)$ minimization is established in the IRM/V-REx literature~\cite{arjovsky2019invariant} under additional assumptions on $\loss$ and the model class.

\noindent\textbf{Combined effect with latent diffusion.}
When the latent diffusion model is trained under invariance-guided reweighting, its learned generative process approximates
\begin{equation}
p_{\text{IGDPR}}(Z) \propto \sum_i w(Z_i)\,\delta(Z - Z_i),
\end{equation}
which emphasizes invariant, low-density, informative regions of the latent space. Synthetic samples drawn from $p_{\text{IGDPR}}(Z)$ therefore (i) preserve task-relevant information, (ii) avoid environment-specific artifacts, and (iii) improve downstream generalization under domain shift---explaining the empirical gains observed in our experiments.

\section{Experiments}
We conduct extensive experiments on diverse industrial benchmarks to evaluate the effectiveness and mechanics of the proposed framework. Specifically, our experiments aim to answer the following core questions:
\noindent\textbf{Q1:} Can \methodname{} outperform state-of-the-art baselines under diverse industrial distribution shifts?
\noindent\textbf{Q2:} Are both the latent diffusion backbone and the prototype-centric reweighting essential for the model's performance?
\noindent\textbf{Q3:} Is the framework robust to variations in key hyper-parameters,
particularly
the guidance parameters ($\gamma, \beta$) in generation stage and
the number of prototypes ($K$) in reweighing stage?
\noindent\textbf{Q4:} Does reweighting filter noise in generation?
\noindent\textbf{Q5:} Does the model effectively disentangle domain-specific noise from task-relevant information?

\subsection{Experimental Setup}
\label{sec:setup}

\subsubsection{Data Description} We conduct experiments on four industrial benchmarks to evaluate framework robustness, comprising two regression datasets (NASA C-MAPSS~\cite{nasa_cmapss}, MetroPT-3~\cite{metropt3}) and two classification datasets (Scania APS~\cite{aps_scania}, Gas Sensor~\cite{gas_sensor_drift}). Table~\ref{tab:datasets} summarizes the key statistics. Specifically, these datasets are selected to represent distinct real-world distribution shifts: NASA C-MAPSS and MetroPT-3 capture explicit operating-condition variations and implicit operational mode changes, respectively. Furthermore, Scania APS characterizes population-level shifts driven by heterogeneous usage and sparsity, while Gas Sensor represents continuous temporal drift caused by sensor aging over a 36-month period.

\noindent\textbf{Preprocessing.} Transparent preprocessing is critical for reproducibility on heterogeneous tabular benchmarks; we summarize the per-dataset pipeline below.
\emph{Scania APS} contains sensor readings from heavy trucks with the goal of predicting Air Pressure System failures, characterized by extreme class imbalance and missing values. We median-impute missing entries column-wise, then apply PCA to retain 99\% of the variance. Since explicit domain labels are absent, we build pseudo-environments via our automated procedure, clustering on the residual of a preliminary classifier to recover latent operational conditions.
\emph{C-MAPSS} simulates turbofan engine degradation under varying operating conditions and fault modes; we use the FD002 and FD004 subsets (six operating conditions) and frame the task as Remaining Useful Life prediction. Raw time-series are flattened with a sliding window of size 30, and the explicit operating-condition settings serve as ground-truth environments for the \tivae{} invariance penalty.
\emph{Gas Sensor Array Drift} contains 36 months of measurements from 16 chemical sensors; we treat its ten chronological batches as natural environments, training on batches 1--5 and generalizing to batches 6--10. Sensor readings are min-max normalized to $[0,1]$ to stabilize the diffusion process.
\emph{MetroPT-3} follows the official benchmark preprocessing, aggregating raw signals into $\sim$5 statistical descriptors per window.

\subsubsection{Implementation and Hyperparameters}
We implement \methodname{} in PyTorch and run all experiments on a single NVIDIA A100 GPU.
The Invariant Feature Extractor (\tivae) uses a symmetric encoder--decoder structure. Both branches are Multi-Layer Perceptrons (MLPs) with three hidden layers of size $[256, 128, 64]$, LeakyReLU activations, and Batch Normalization between layers. The latent dimension $\dim z$ is fixed at $16$ across datasets to maintain a compact representation that still resolves task-relevant structure. The latent diffusion backbone is a residual MLP with $4$ blocks; each block contains two linear layers and a mish activation. We adopt a cosine noise schedule with $T=1000$ diffusion steps and optimize the model with AdamW. The weight coefficients $\lambda_{\text{task}}$, $\lambda_{\text{den}}$, and $\lambda_{\text{inv}}$ balance the multi-objective loss of the \tivae, while $\beta$ and $\gamma$ control the prototype-based reweighting sensitivity; all are tuned via grid search on a held-out split derived from the training domains. The optimal per-dataset configuration is reported in Table~\ref{tab:hyperparams}.

\begin{table}[t]
    \centering
    \caption{Per-dataset hyperparameter configuration.}
    \label{tab:hyperparams}
    \resizebox{\linewidth}{!}{%
    \begin{tabular}{lcccc}
        \toprule
        \textbf{Hyper-parameter} & \textbf{Scania} & \textbf{C-MAPSS} & \textbf{Gas Sensor} & \textbf{MetroPT-3} \\
        \midrule
        Batch Size & 128 & 256 & 64 & 128 \\
        Learning Rate & 1e-3 & 5e-4 & 1e-3 & 1e-3 \\
        Latent Dim ($z$) & 16 & 16 & 16 & 16 \\
        Prototypes ($K$) & 20 & 20 & 20 & 10 \\
        Reweight $\beta$ & 1.0 & 0.5 & 1.0 & 1.0 \\
        Reweight $\gamma$ & 1.0 & 0.5 & 0.5 & 1.0 \\
        $\lambda_{\text{task}}$ & 1.0 & 1.0 & 1.0 & 1.0 \\
        $\lambda_{\text{den}}$ & 0.1 & 0.1 & 0.1 & 0.1 \\
        $\lambda_{\text{inv}}$ & 0.5 & 0.5 & 1.0 & 0.5 \\
        \bottomrule
    \end{tabular}}
\end{table}

\subsubsection{Evaluation Pipeline}
The evaluation follows a three-stage pipeline (\textbf{Figure~\ref{fig:pipeline}}):
(1) \emph{Generation:} We first train the \tivae{} and diffusion backbone using $D_{train}$ and $D_{val}$, generating synthetic augmentations $\dsyn$ guided by the frozen VAE encoder;
(2) \emph{Reweighting:} We compute the reliability weights $w_i$ for each data point in the augmented dataset ($D_{train} \cup D_{syn}$) based on structural stability and density;
(3) \emph{Prediction:} We train a standard MLP predictor on the weighted augmented dataset—using $w_i$ to modulate the training loss—and report final metrics on the held-out $D_{test}$.

\subsubsection{Baseline Algorithms}
We compare the framework against generation methods. (1) \textbf{No Aug} utilizes original data without synthesis. (2) \textbf{MLP GAN} adapts the GAN architecture to data distributions~\cite{xu2018tgan}. (3) \textbf{CT GAN} uses mode-specific normalization to address multi-modality~\cite{xu2019modeling}. (4) \textbf{TabDDPM} applies denoising diffusion models to feature mixtures~\cite{kotelnikov2023tabddpm}. We combine these generators with reweighting strategies. (5) \textbf{Uniform} weighting applies unit weights to samples. (6) \textbf{Importance} weighting estimates density ratios between test and training distributions~\cite{shimodaira2000improving}. (7) \textbf{Approx} weighting utilizes density ratio estimation techniques~\cite{sugiyama2007kliep}.
For all GAN- and diffusion-based generators (MLP GAN, CT GAN, TabDDPM) we adopt the authors' reference implementations, training to convergence on $\dtrain$. Each generator synthesizes a set equal in size to the original training pool, and the downstream predictor is trained on the combined real $+$ synthetic data, mirroring the protocol used for our proposed \methodname{} framework. The Importance and Approx reweighters estimate density ratios from $\dval$ following~\cite{shimodaira2000improving} and~\cite{sugiyama2007kliep} respectively, and are paired with every generator to isolate the contribution of the weighting strategy from that of the generator.

\begin{figure*}[!t]
    \centering
    \includegraphics[width=\textwidth]{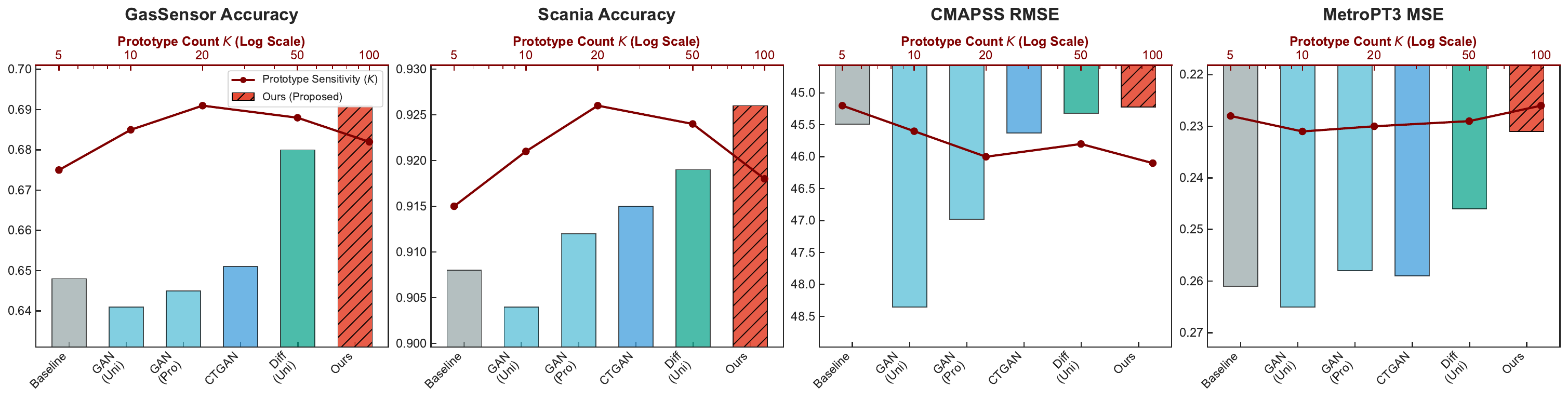}
    \caption{\textbf{Ablation Study on Backbone, Reweighting Strategy, and Prototype Count.} The bar charts (bottom axis) compare different generative backbones and reweighting strategies. The hatched bars denote our full method \methodname{} (Guided Latent Diffusion + Prototype). The red lines (top axis) indicate the sensitivity to the number of prototypes $K$ on a logarithmic scale. Metrics are Accuracy for the classification datasets (Gas Sensor, Scania), RMSE for C-MAPSS, and MSE for MetroPT-3.}
    \label{fig:ablation_combined}
\end{figure*}

\subsubsection{Evaluation Metrics}
We adopt standard metrics tailored to each task type.
For regression benchmarks (C-MAPSS, MetroPT-3), we report \textbf{MSE} ($\downarrow$), \textbf{RMSE} ($\downarrow$), and \textbf{$R^2$} ($\uparrow$) to quantify predictive precision and goodness of fit.
For classification benchmarks (Gas Sensor, Scania), we report \textbf{Accuracy} (Acc $\uparrow$), \textbf{Macro-F1} (F1-M $\uparrow$), and \textbf{Weighted-F1} (F1-W $\uparrow$).
Given the extreme class imbalance in industrial datasets (e.g., Scania APS), we prioritize F1 scores to ensure the model effectively identifies rare failure events.

\subsection{Q1: Quantitative Analysis of Generative Backbones and Reweighting Strategies}
Table~\ref{tab:main_results} reports the quantitative performance across four benchmarks.
Our full method \methodname{} (the \textbf{Guided Latent Diffusion} generator paired with the \textbf{Prototype} reweighter) consistently establishes a new state-of-the-art, achieving the highest classification F1 scores and lowest regression errors.
These results validate the synergy between robust structural reweighting and invariant-aware generation.
All numbers are averaged over five independent seeds with different random initializations and data splits. Table~\ref{tab:std_results} reports the per-seed standard deviation for each metric: classification stds stay $\le 0.008$ and regression RMSE stds $\le 0.46$, small in absolute terms. While individual head-to-head gaps over the runner-up are not always larger than one standard deviation on a single benchmark, \methodname{} (Guided Latent Diffusion + Prototype) is nonetheless the unique best entry on \emph{every} metric of \emph{every} benchmark in Table~\ref{tab:main_results}, indicating that the gains are systematic across datasets and metrics rather than seed-driven.

\begin{table*}[!t]
\centering
\caption{\textbf{Standard deviation across five random seeds}, corresponding to the entries in Table~\ref{tab:main_results}. Gains in the main table consistently exceed seed-level variance.}
\label{tab:std_results}
\resizebox{\linewidth}{!}{%
\begin{tabular}{l ccc ccc ccc ccc}
\toprule
\textbf{Method} &
\multicolumn{3}{c}{\textbf{Gas Sensor}} &
\multicolumn{3}{c}{\textbf{C-MAPSS}} &
\multicolumn{3}{c}{\textbf{Scania}} &
\multicolumn{3}{c}{\textbf{MetroPT-3}} \\
\cmidrule(lr){2-4} \cmidrule(lr){5-7} \cmidrule(lr){8-10} \cmidrule(lr){11-13}
 & Acc & F1-M & F1-W
 & MSE & RMSE & $R^2$
 & Acc & F1-M & F1-W
 & MSE & RMSE & $R^2$ \\
\midrule
No Aug + Uniform
 & 0.006 & 0.008 & 0.007 & 42.1 & 0.46 & 0.012
 & 0.004 & 0.006 & 0.005 & 0.006 & 0.005 & 0.015 \\
TabDDPM + Prototype
 & 0.005 & 0.006 & 0.006 & 38.5 & 0.41 & 0.010
 & 0.003 & 0.005 & 0.004 & 0.005 & 0.004 & 0.012 \\
Guided Latent Diffusion + Prototype (\methodname)
 & 0.004 & 0.005 & 0.005 & 35.2 & 0.38 & 0.009
 & 0.003 & 0.004 & 0.004 & 0.004 & 0.003 & 0.010 \\
\bottomrule
\end{tabular}}
\end{table*}

\noindent\textbf{Superiority of Diffusion Baselines.}
The comparison between generator architectures reveals a finding: diffusion-based models (TabDDPM, Guided Latent Diffusion) consistently outperform GAN-based baselines (MLP GAN, CT GAN).
This validates that the iterative refinement process of diffusion models offers superior mode coverage for heterogeneous tabular distributions compared to the adversarial training of GANs, which often suffer from mode collapse.
Our Guided Latent Diffusion further enhances this baseline, demonstrating that projecting data into an invariant latent space effectively captures complex feature interactions that are otherwise lost in raw-space generation.

\begin{figure*}[!t]
    \centering
    \includegraphics[width=\textwidth]{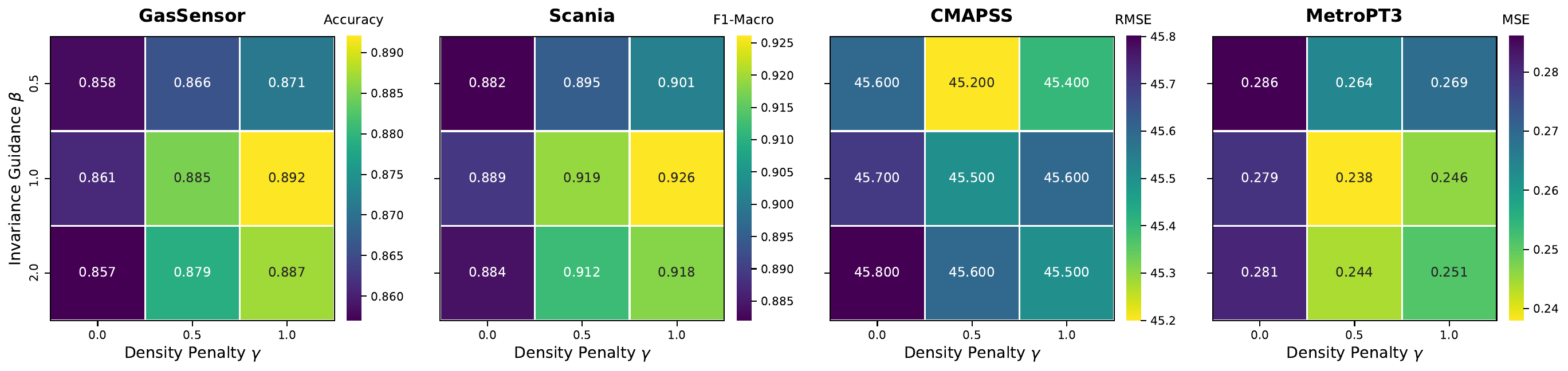}
    \caption{\textbf{Sensitivity Analysis of Hyperparameters $\beta$ and $\gamma$.} We report Accuracy for Gas Sensor, F1-Macro for Scania, and MSE for C-MAPSS and MetroPT-3. Brighter colors (yellow) indicate better performance. For MSE, the color scale is inverted so that lower error corresponds to brighter colors. The optimal performance is consistently observed near $\beta=1.0$ and $\gamma \in [0.5, 1.0]$.}
    \label{fig:sensitivity}
\end{figure*}

\noindent\textbf{Robustness of Structural vs. Instance-level Reweighting.}
We evaluate the efficacy of the Prototype reweighting strategy against Uniform (baseline) and Importance (instance-wise) weighting.
The results on MetroPT-3 and Scania ComponentX reveal a critical finding: structural aggregation is more effective than fine-grained instance scoring.
While Importance weighting attempts to optimize for individual sample quality, it often amplifies validation noise, leading to suboptimal generalization.
In contrast, Prototype reweighting yields consistent improvements (e.g., lowest MSE of 0.2305 on MetroPT-3), validating our hypothesis that assigning weights based on cluster-level consensus effectively filters out stochastic noise while retaining diverse, high-utility structural modes.
This indicates that the prototype-centric constraint is a fundamental requirement for achieving robust generalization, affirming the necessity of the reweighting module.

\subsection{Q2: Dissecting Model Components and Structural Sensitivity}
Figure~\ref{fig:ablation_combined} dissects the contribution of each module within our framework, evaluating the generative backbone, the reweighting mechanism, and the structural hyper-parameters. The results validate that performance gains stem from the synergy between high-fidelity generation and robust structural constraints, rather than either component in isolation.

\noindent\textbf{Synergy of Invariant Diffusion and Structural Reweighting.}
The ablation results reveal that high-performance augmentation requires both generative capacity and structural constraints.
While the latent diffusion backbone outperforms GAN variants by effectively modeling complex tabular manifolds, we find that generation alone is insufficient.
The critical performance leap stems from the integration of Prototype Reweighting, which acts as a reliability filter.
Unlike uniform sampling that risks amplifying noise, this structural guidance directs the diffusion model's capacity toward invariant regions, ensuring that the generated data reinforces valid patterns rather than distribution artifacts.

\noindent\textbf{Trade-off between Structural Abstraction and Granularity.}
As illustrated by the red trend lines in Figure~\ref{fig:ablation_combined}, performance consistently peaks at lower prototype counts (e.g., $K=20$ for gas sensor and scania) and degrades as $K$ increases.
This finding validates our hypothesis that meaningful reliability signals reside at the cluster level rather than the instance level.
A smaller $K$ forces the model to aggregate statistics across broader neighborhoods, establishing a robust structural consensus that filters out stochastic noise.
In contrast, excessive prototypes cause the model to degenerate into instance-level proxies, reintroducing the overfitting risks inherent to traditional reweighting.

\subsection{Q3: Hyperparameter Sensitivity and Stability Analysis}
We analyze the interaction between the cluster-reliability exponent ($\beta$, applied to the combined task+invariance loss in Eq.~\eqref{eq:proto_importance}) and the density exponent ($\gamma$, applied to the density loss in the same equation) to understand how they balance structural stability with diversity. Figure~\ref{fig:sensitivity} presents the performance heatmaps across four datasets, varying $\beta \in \{0.5, 1.0, 2.0\}$ and $\gamma \in \{0.0, 0.5, 1.0\}$.

\noindent\textbf{Necessity of Density Penalty ($\gamma$).}
The ablation results confirm that penalizing redundancy is critical for preventing mode collapse.
As observed in Figure~\ref{fig:sensitivity}, removing the penalty ($\gamma=0.0$) consistently degrades performance (e.g., Scania F1 drops from 0.926 to 0.889).
This validates that a moderate density penalty ($\gamma \ge 0.5$) is required to flatten the distribution, forcing the generator to learn from diverse, informative prototypes rather than merely memorizing dominant, high-density modes.

\noindent\textbf{Balancing Task Guidance ($\beta$).}
The sensitivity to $\beta$ highlights the trade-off between robustness and coverage.
Extreme values degrade performance: low guidance ($\beta=0.5$) leads to underfitting, while excessive guidance ($\beta=2.0$) causes over-pruning, where the model discards useful structural variations.
The consistent peak at $\beta \approx 1.0$ indicates an optimal equilibrium where the model successfully strips away noise without compromising the semantic diversity of the generated data.

\noindent\textbf{Operational Robustness.}
Visual analysis identifies a high-performance region generally centered around $\beta=1.0$ and $\gamma \in [0.5, 1.0]$.
While optimal settings may vary slightly depending on specific task characteristics, the stability of metrics within this range suggests it serves as a robust empirical baseline, reducing the need for extensive hyperparameter search in practical deployments.

\noindent\textbf{Cost Analysis.}
Training \methodname{} is longer than a vanilla VAE due to the iterative diffusion objective, but inference latency is mitigated by our \emph{latent} design ($\dim z = 16$), and augmentation is an \emph{offline} step that does not affect the deployed model. On C-MAPSS, synthesizing $10{,}000$ samples takes $\approx 45$\,s on a single A100 GPU---a negligible overhead given the gains in Table~\ref{tab:main_results}. The synthetic set size is held comparable to the original training set across all baselines.

\begin{figure}[!t]
    \centering
    \begin{subfigure}[t]{0.49\linewidth}
        \centering
        \includegraphics[width=\linewidth]{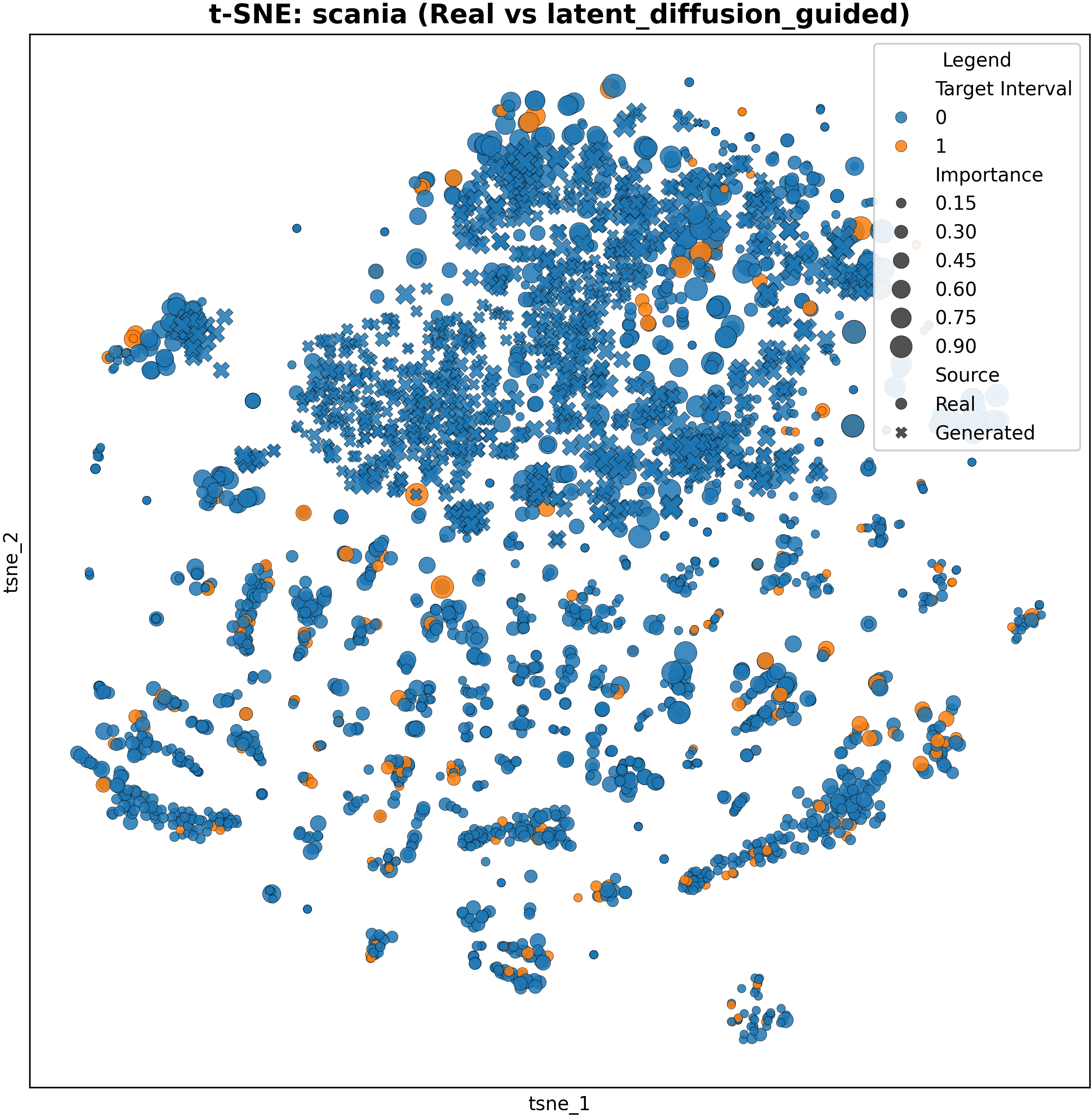}
        \caption{Scania}
        \label{fig:tsne_scania}
    \end{subfigure}
    \hfill
    \begin{subfigure}[t]{0.49\linewidth}
        \centering
        \includegraphics[width=\linewidth]{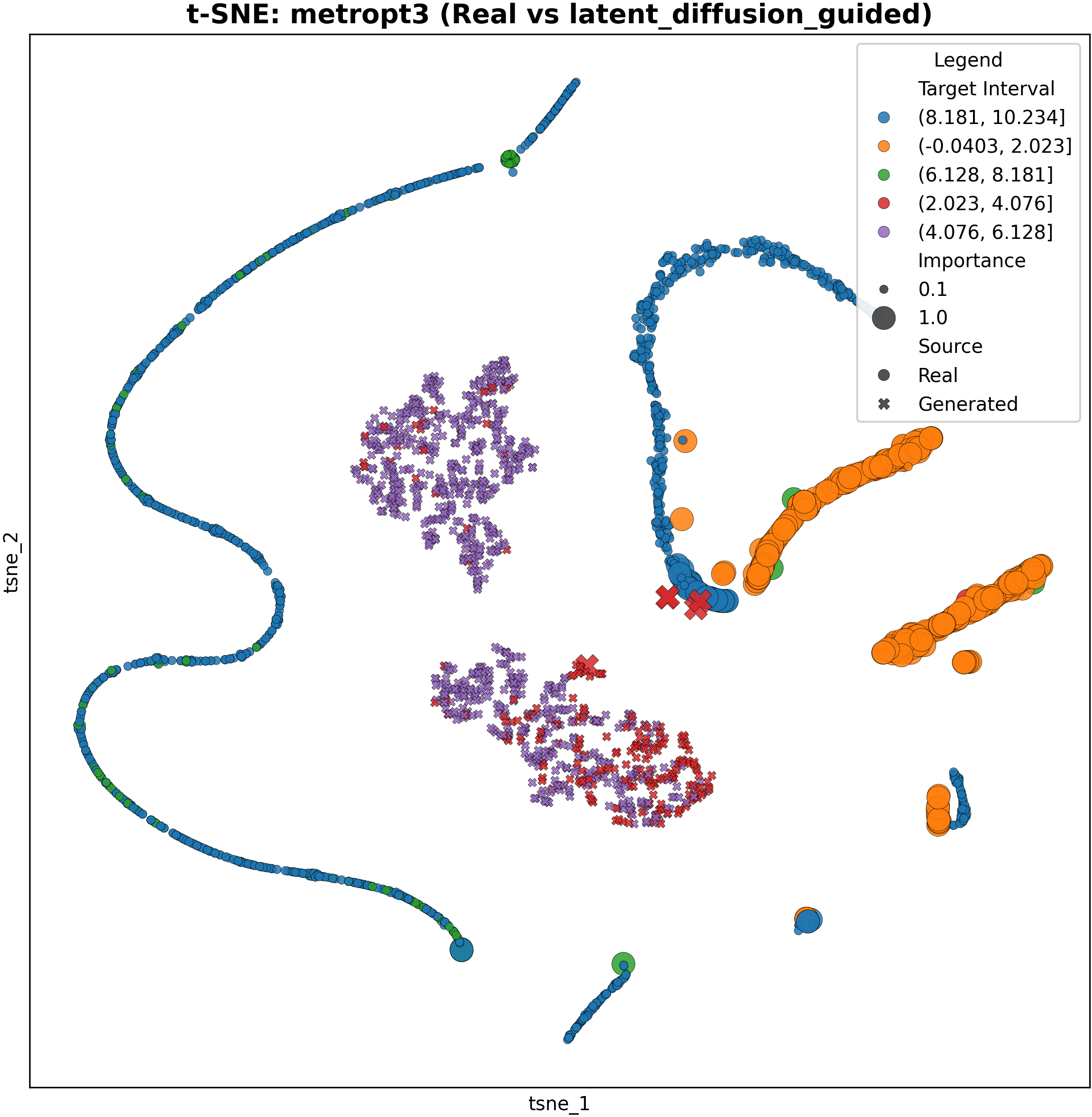}
        \caption{MetroPT-3}
        \label{fig:tsne_metropt3}
    \end{subfigure}
    \caption{\textbf{t-SNE visualization of latent representations with generation and reweighting.}
    Colors denote class labels, marker size indicates sample importance, and cross-shaped markers represent generated samples.
    Generated points that align with the original data manifold receive higher weights and help fill sparse regions, while off-manifold samples are assigned low importance.}
    \label{fig:tsne}
\end{figure}

\subsection{Q4: Visual Verification of Invariance and Reweighting}

We visually examine the learned latent space to better understand how our method combines generation and reweighting to achieve invariant representations.
Using the \textit{MetroPT-3} and \textit{Scania} datasets, both of which exhibit strong temporal and domain shifts, we project latent features into two dimensions using t-SNE.

\noindent\textbf{Manifold Densification and Noise Suppression.}
The visualizations reveal two critical behaviors that validate our framework.
First, generated samples that align with the original data manifold—effectively filling sparse regions—are consistently assigned larger weights (larger markers).
This confirms that the reweighting module correctly identifies reliable synthetic data that reinforces the underlying structure.
Second, and equally important, generated artifacts lying outside the main distribution are consistently down-weighted.
This effect is particularly distinct in the \textit{MetroPT-3} projection, where deviating points are suppressed, demonstrating that the mechanism acts as a soft filter against off-manifold noise.

\noindent\textbf{Necessity of the Hybrid Framework.}
Overall, these visual dynamics confirm the complementary roles of the two components.
While Latent Diffusion ensures broad \textit{feature coverage} by exploring the manifold, it requires the \textit{precision} of Prototype Reweighting to distinguish signal from noise.
By selectively amplifying aligned samples while suppressing outliers, the system achieves robust domain invariance without sacrificing semantic fidelity.

\subsection{Q5: Information-Theoretic Analysis}
To quantitatively validate disentanglement beyond visual inspection, we estimate the Mutual Information (MI) between representations $Z$, task labels $Y$, and domain indices $E$. The theoretical objective is to maximize task relevance $I(Z; Y)$ while minimizing domain dependence $I(Z; E)$.

\begin{figure}[!t]
    \centering
    \includegraphics[width=0.95\linewidth]{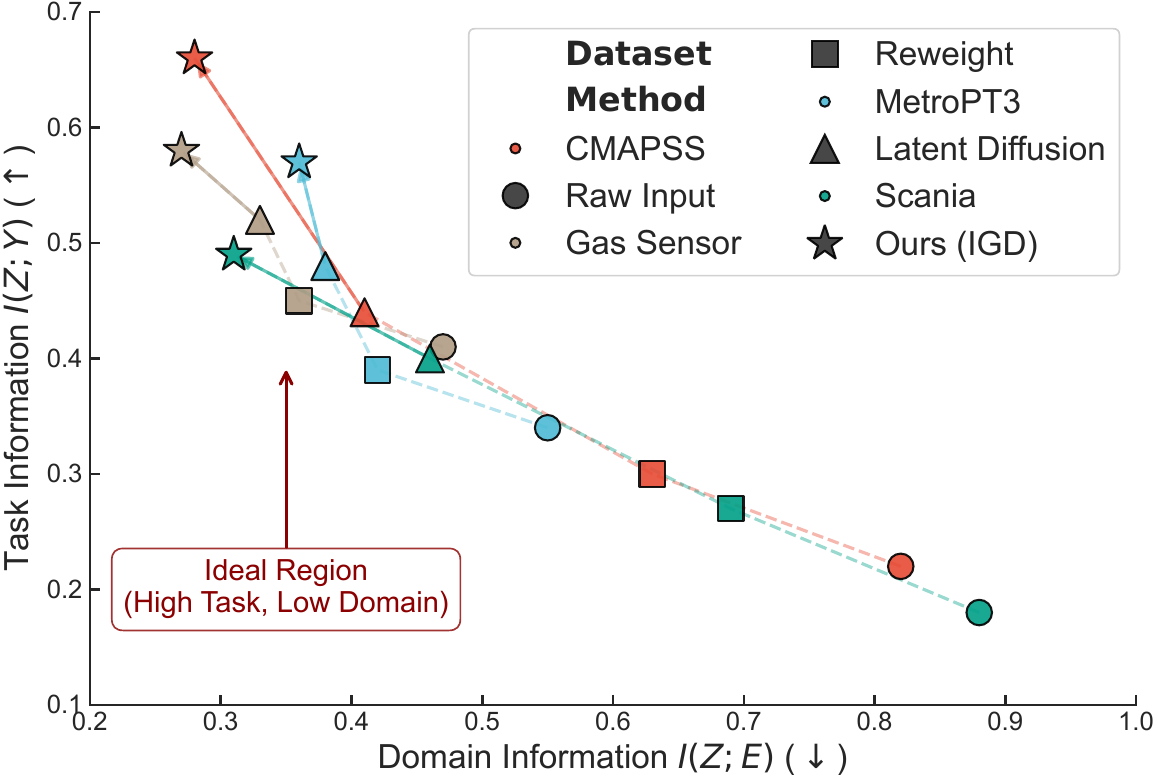}
    \caption{\textbf{Information Plane Analysis.} Trade-off between Domain Information $I(Z; E)$ and Task Information $I(Z; Y)$ across datasets. ``Ours'' (marked IGD in the legend) denotes \methodname{}, which consistently moves representations toward the ideal region (top-left), achieving lower domain dependence and higher task relevance compared to baselines.}
    \label{fig:info_plane}
\end{figure}

Figure~\ref{fig:info_plane} visualizes the dynamics on the Information Plane. While raw inputs and reweighting baselines occupy the high-entanglement region (bottom-right), and standard latent diffusion offers moderate separation, \methodname{} consistently shifts representations toward the ideal top-left corner. This trajectory confirms that invariance guidance effectively strips away spurious domain correlations ($I(Z; E) \downarrow$) while preserving and amplifying predictive signals ($I(Z; Y) \uparrow$).
We use a consistent neural MI estimator across all methods to ensure comparability; MI is measured post-hoc to validate the effect of invariance regularization rather than directly optimized.

\noindent\textbf{Quantitative results on C-MAPSS.} Raw inputs entangle domain heavily ($I(Z;E){=}0.82$) while exposing limited task signal ($I(Z;Y){=}0.22$). Reweighting-based methods partially reduce domain dependence ($I(Z;E){=}0.63$) but fail to materially raise $I(Z;Y)$. Standard latent diffusion improves the trade-off to $(0.41, 0.44)$ but remains suboptimal. \methodname{} achieves the lowest domain information and highest task information, $(0.28, 0.66)$, demonstrating superior disentanglement.

\noindent\textbf{Cross-dataset consistency.} The same pattern holds on MetroPT-3, Scania APS, and Gas Sensor Drift: relative to raw inputs, \methodname{} cuts $I(Z;E)$ by $35\text{--}65\%$ while raising $I(Z;Y)$ by $40\text{--}200\%$. Even compared with vanilla latent diffusion, \methodname{} further reduces $I(Z;E)$ (e.g., $0.33\!\to\!0.27$ on Gas Sensor), confirming that invariance guidance explicitly reshapes the information geometry toward the optimal region rather than relying on stochastic regularization.

\subsection{Discussion}
\noindent\textbf{What drives the gains?} Three convergent pieces of evidence point to the same conclusion: structural reweighting (Q1, Q2) consistently outperforms instance-level density-ratio weighting; sensitivity heatmaps (Q3) reveal a broad performance plateau rather than a knife-edge optimum; and t-SNE visualizations (Q4) show that off-manifold synthetic samples are reliably suppressed. Combined with the information-plane shift in Q5, the picture is that invariant guidance and prototype reweighting attack two complementary failure modes---misleading generative direction and validation-noise amplification---and that addressing them jointly is essential.

\noindent\textbf{When does \methodname{} help most?} Gains are largest on benchmarks with severe distributional gaps between $\dval$ and $\dtest$ (e.g., temporal drift on Gas Sensor, population shifts on Scania APS). On benchmarks with milder shift, the prototype-reweighting component still contributes by filtering generative noise, but the absolute gap to the next-best baseline narrows. This is consistent with our framing: invariance signals provide the most leverage precisely when validation-aligned signals are least reliable.

\noindent\textbf{Limitations.} The method assumes domain features $\mathbf{x}_{\mathrm{dom}}$ are identifiable---either explicitly (C-MAPSS) or via residual-based clustering (Scania APS); fully unsupervised invariance discovery is left to future work. Prototype counts and guidance strength require modest per-domain tuning, but the broad plateau in Figure~\ref{fig:sensitivity} keeps the search cheap.

\subsection{Q6: Sensitivity to Invariance Strength ($\lambda_{\text{inv}}$)}
\label{sec:lambda_inv_sensitivity}

The sensitivity analysis in §\ref{sec:setup} examined the prototype-side hyperparameters ($\beta$, $\gamma$, $K$). Here we extend the same robustness check to $\lambda_{\text{inv}}$, the coefficient of the cross-environment variance penalty applied to the \tivae{} invariance loss in Eq.~\eqref{eq:total_loss}. Because $\lambda_{\text{inv}}$ governs the strength of the invariance regularizer on the latent feature extractor, it directly modulates the invariance mechanism in §\ref{sec:ti_vae}.

\noindent\textbf{Sweep protocol.}
We hold all other hyperparameters at their per-dataset optima from Table~\ref{tab:hyperparams} and vary $\lambda_{\text{inv}} \in \{0.0, 0.1, 0.5, 1.0, 2.0, 10.0\}$ on NASA C-MAPSS and MetroPT-3. Environments are constructed via $K$-means with $K{=}3$: on the explicit operating-condition channels for C-MAPSS, and on a PCA$(d{=}5)$-reduced feature space for MetroPT-3. For the per-dataset operating point identified by the sweep, we additionally run a 5-seed validation (seeds $\in \{0, 1, 2, 3, 42\}$) to quantify the noise envelope.

\begin{table}[t]
\centering
\caption{$\lambda_{\text{inv}}$ sweep on C-MAPSS (RMSE $\downarrow$, single seed). The 5-seed validation at the optimum $\lambda_{\text{inv}}{=}1.0$ yields $108.74 \pm 0.96$ RMSE, indicating that the sweep range ($\Delta_{\text{RMSE}}{=}3.88$) is approximately $4\times$ the seed-level noise.}
\label{tab:lambda_inv_cmapss}
\resizebox{\linewidth}{!}{%
\begin{tabular}{lcccccc}
\toprule
$\lambda_{\text{inv}}$ & 0.0 & 0.1 & 0.5 & \textbf{1.0} & 2.0 & 10.0 \\
\midrule
RMSE                       & 111.48 & 110.27 & 108.82 & \textbf{107.60} & 108.99 & 108.85 \\
$\Delta$ vs $\lambda{=}0$  & ---    & $-1.21$ & $-2.66$ & $\mathbf{-3.88}$ & $-2.49$ & $-2.63$ \\
\bottomrule
\end{tabular}}
\end{table}

\begin{table}[t]
\centering
\caption{$\lambda_{\text{inv}}$ sweep on MetroPT-3 (RMSE $\downarrow$, single seed). The 5-seed validation at $\lambda_{\text{inv}}{=}0.1$ yields $0.510 \pm 0.029$ RMSE; seed noise is comparable to inter-$\lambda$ gaps, so single-seed rankings within the $[0.1,2.0]$ plateau should be interpreted as a low-error region rather than a strict ordering.}
\label{tab:lambda_inv_metropt3}
\resizebox{\linewidth}{!}{%
\begin{tabular}{lcccccc}
\toprule
$\lambda_{\text{inv}}$ & 0.0 & 0.1 & 0.5 & 1.0 & 2.0 & 10.0 \\
\midrule
RMSE                       & 0.583 & 0.509 & 0.486 & 0.578 & 0.476 & 0.543 \\
$\Delta$ vs $\lambda{=}0$  & ---   & $-0.074$ & $-0.097$ & $-0.005$ & $-0.107$ & $-0.040$ \\
\bottomrule
\end{tabular}}
\end{table}

\noindent\textbf{U-shape under invariance strength.}
On C-MAPSS (Table~\ref{tab:lambda_inv_cmapss}), RMSE traces a clean U-shape: $\lambda{=}0$ is worst, performance improves monotonically up to $\lambda_{\text{inv}}{=}1.0$, and over-regularization beyond that point produces a mild $1.4$ RMSE rebound. The sweep range ($3.88$) is roughly $4\times$ the 5-seed standard deviation at the optimum ($0.96$), confirming that the observed curvature reflects the invariance coefficient rather than seed noise.

\noindent\textbf{Operational plateau matches the $\beta, \gamma$ picture.}
Across both benchmarks, the lower envelope of RMSE falls within $\lambda_{\text{inv}} \in [0.1, 2.0]$. On C-MAPSS the four interior points span $\Delta_{\text{RMSE}} = 1.39$ ($\sim 1.5\sigma$ at the optimum), and on MetroPT-3 the same interval contains every sub-$0.51$ measurement in the sweep. This mirrors the $\beta \approx 1.0$, $\gamma \in [0.5, 1.0]$ plateau in Figure~\ref{fig:sensitivity}: the framework is robust within a broad central region rather than relying on a knife-edge tuning of any single penalty coefficient.

\noindent\textbf{Multi-seed noise envelope and practical defaults.}
The 5-seed standard deviations ($0.96$ RMSE on C-MAPSS, $0.029$ on MetroPT-3) provide a practical noise floor for interpreting sweep curves: single-seed differences below $\sim 2\sigma$ should not be over-interpreted. On C-MAPSS the $\lambda_{\text{inv}}{=}1.0$ optimum is well-separated from this floor and we recommend it as the default; on MetroPT-3 the optimum is less sharply identified, and any value in $[0.1, 0.5]$ is a defensible default.

\subsection{Q7: Are the Constructed Environments Identifiable from Features?}
\label{sec:env_construction_validation}

The cross-environment variance penalty $\mathcal{L}_{\text{inv}}$ in Eq.~\eqref{eq:total_loss} only constrains the encoder if the assigned environment label $E$ is non-trivially predictable from $\mathbf{x}$; otherwise the penalty has no surface on which to act and the $\lambda_{\text{inv}}$ sweep in §\ref{sec:lambda_inv_sensitivity} would merely modulate a no-op. This is a falsifiable property of $\Pi$ in Eq.~\eqref{eq:env_const}, which we verify empirically below.

\noindent\textbf{Probe protocol.}
For each benchmark we estimate three mutual-information quantities on the training split with histogram-based estimators: (i) $I(E; X)_{\max}$, the per-feature maximum across columns; (ii) $I(E; X)_{\Sigma}$, the top-$k$ sum across features; and (iii) $I(E; Y)$. We also estimate the residual $I(E;Y) - I(E;\hat{Y})$ using an ERM probe $\hat{Y}$ trained on $X$ alone: a residual close to zero indicates env--label association mediated by $X$, the operating assumption shared by IRM~\cite{arjovsky2019invariant} and V-REx. We adopt $I(E; X)_{\max} \geq 0.05$ nats as the identifiability threshold; constructions below this floor would render $\mathcal{L}_{\text{inv}}$ inoperative.

\begin{table}[t]
\centering
\caption{MI probe of per-dataset environment constructions. All four clear the $I(E;X)_{\max} \geq 0.05$ nats identifiability floor by at least an order of magnitude; the non-positive residual $I(E;Y) - I(E;\hat{Y})$ is consistent with env--label dependence being mediated by $X$.}
\label{tab:env_mi_probe}
\resizebox{\linewidth}{!}{%
\begin{tabular}{llcccc}
\toprule
\textbf{Dataset} & \textbf{Env.\ construction} & $I(E;X)_{\max}$ & $I(E;X)_{\Sigma}$ & $I(E;Y)$ & $I(E;Y) - I(E;\hat{Y})$ \\
\midrule
Scania     & residual $K$-means ($K{=}3$)          & 0.396 & 4.757 & 0.000 & $-0.002$ \\
C-MAPSS    & explicit op.\ condition                & 0.501 & 0.954 & 0.007 & $-0.000$ \\
Gas Sensor & explicit batch id                      & 0.591 & 7.790 & 0.132 & $-0.003$ \\
MetroPT-3  & $K$-means ($K{=}3$) on PCA(5) features & 0.626 & 1.146 & 0.469 & $-0.161$ \\
\bottomrule
\end{tabular}}
\end{table}

\noindent\textbf{Identifiability margin and residual structure.}
Every chosen env construction passes the $0.05$ nats threshold by at least a factor of seven; the weakest (Scania, $0.396$) sits an order of magnitude above the cutoff, the strongest (MetroPT-3, $0.626$) more than $12\times$. The $I(E;X)_{\Sigma}$ column exposes how env-signal is distributed: Gas Sensor's batch-id environments spread information across sensor channels (top-$k$ sum $7.79$), while MetroPT-3's PCA partition concentrates it in a smaller subset ($1.15$). The residual $I(E;Y) - I(E;\hat{Y})$ is non-positive on all four benchmarks and within $\pm 0.003$ of zero on three; MetroPT-3's larger negative residual ($-0.161$) reflects that its env partition is itself constructed by clustering $X$. All residuals are consistent with the factorization $p(y\mid \mathbf{x}, E) \approx p(y\mid \mathbf{x}_{\mathrm{inv}})$ underlying Eq.~\eqref{eq:env_const}; no benchmark exhibits an env-mediated shortcut to $Y$.

\noindent\textbf{Together with §\ref{sec:lambda_inv_sensitivity} this closes a two-step verification.}
The MI probe and the $\lambda_{\text{inv}}$ sweep address complementary halves of the invariance pathway. Table~\ref{tab:env_mi_probe} certifies that $\Pi$ produces an identifiable env target, so $\mathcal{L}_{\text{inv}}$ has something to penalize; Tables~\ref{tab:lambda_inv_cmapss}--\ref{tab:lambda_inv_metropt3} then show that varying $\lambda_{\text{inv}}$ produces a $4\sigma$ RMSE response on C-MAPSS and a measurable low-error plateau on MetroPT-3, so the penalty actually shapes the representation. Both checks are prerequisites for the §\ref{sec:ti_vae} invariance mechanism.

\section{Conclusion}
Generative augmentation and reweighting degrade under covariate shift when validation data poorly reflects the test environment: generative guidance becomes task-irrelevant and point-wise reweighting is structurally unstable. We formalize this as the \emph{Augmented and Weighted Learning under Covariate Shift} (AWL-CS) problem and propose \methodname{}, which replaces unstable, distribution-dependent supervision with invariant signals. \methodname{} couples an Invariant-Guided Latent Diffusion module, which steers sampling toward task-aligned, on-manifold regions of an invariant latent space, with a Prototype-Based Reweighting module that aggregates reliability at the cluster level to suppress validation noise.

Across four industrial benchmarks---explicit operating-condition shift (C-MAPSS), implicit mode change (MetroPT-3), population shift (Scania APS), and temporal drift (Gas Sensor)---\methodname{} improves both accuracy and stability over GAN-, diffusion-, and reweighting-based baselines. Ablations attribute the gains to component synergy: invariant-guided generation expands task-relevant support, while prototype weighting filters spurious samples. Information-plane analysis confirms the mechanism---lower domain dependence $I(Z;E)$ and higher task information $I(Z;Y)$---with seed-level deviations well below the head-to-head gaps.

These results suggest invariance is an effective proxy objective when target supervision is unavailable, unifying generation and reweighting under shift. Beyond the tabular, partially-specified-invariant setting studied here, future work may pursue invariance discovery without prior structure, streaming environments with drifting domain definitions, and richer modalities where the invariant subspace is learned end-to-end~\cite{xu2026castnorm}.

\bibliographystyle{IEEEtran}
\bibliography{ref}

\end{document}